\PassOptionsToPackage{numbers, compress}{natbib}
\documentclass{article}

 \usepackage{realbench_arxiv}
 \paperbrand{RealBench}
 \paperkicker{Operational Weather Forecasting Benchmark}
 \preprintnotice{arXiv Preprint}
 \papercontact{\textbf{Corresponding:} \textcolor{RBBlue}{\texttt{songguo@ust.hk}}; \textcolor{RBBlue}{\texttt{bailei@pjlab.org.cn}}}

\usepackage[utf8]{inputenc} 
\usepackage[T1]{fontenc}    
\usepackage{hyperref}       
\usepackage{url}            
\usepackage{booktabs}       
\usepackage{amsfonts}       
\usepackage{nicefrac}       
\usepackage{microtype}      
\usepackage{xcolor}         

\usepackage{enumitem}
\usepackage{graphicx}
\usepackage{booktabs}
\usepackage{multirow}
\usepackage{amsmath}
\usepackage{booktabs}
\usepackage{array}    
\usepackage{url} 
\makeatletter
\g@addto@macro{\UrlBreaks}{\do\1\do\2\do\3\do\4\do\5\do\6\do\7\do\8\do\9\do\0\do\-\do\_}
\makeatother
\usepackage{ragged2e}

\title{RealBench: Benchmarking Data-Driven Numerical Weather
Forecasting \\ Under Operational Conditions and Extreme Event Challenges}

\author{%
{\large
\textbf{Ruize Li}$^{1,2}$\RBEqual \quad
\textbf{Zhibin Wen}$^{3}$\RBEqual \quad
\textbf{Tao Han}$^{1,4}$\RBEqual\RBLead \quad
\textbf{Hao Chen}$^{1}$ \quad
\textbf{Fenghua Ling}$^{4}$ \quad
\textbf{Wei Zhang}$^{5}$}\\[0.25em]
{\large
\textbf{Song Guo}$^{1}$\RBCorresponding \quad
\textbf{Lei Bai}$^{4}$\RBCorresponding}\\[0.55em]
{\fontsize{9.6}{11.3}\selectfont
$^1$The Hong Kong University of Science and Technology \quad $^2$Nanjing University\\
$^3$Southern University of Science and Technology \quad $^4$Shanghai AI Laboratory\\ 
$^5$Shanghai TechWind Technology Co., Ltd.}\\[0.45em]
{\small
\textbf{*}Equal contribution \quad
\S Corresponding authors \quad
\ensuremath{\dagger}Project lead}
}

\begin{document}

\maketitle

\begin{abstract}

  Accurate evaluation of weather forecasting models is critical for their reliable deployment in real-world applications. 
  However, existing benchmarks predominantly rely on reanalysis products such as ERA5, which are generated through delayed data assimilation and do not reflect the constraints of real-time operational forecasting, thereby resulting in a systematic mismatch between benchmark performance and real-world forecasting. 
  In this work, we introduce RealBench, a next-generation benchmark for AI weather forecasting that emphasizes realistic evaluation under operational conditions. 
  RealBench features a strictly out-of-distribution test set spanning 2025 to eliminate data leakage and capture recent atmospheric regimes. It integrates multiple data sources, including low-latency operational analysis and a large-scale global in-situ observation dataset comprising over 10,000 stations, enabling direct evaluation against real atmospheric measurements. Beyond standard global metrics, RealBench provides a comprehensive evaluation framework for high-impact extreme events, including heatwaves, cold surges, and tropical cyclones, using event-specific metrics that better reflect real-world forecasting priorities. 
  The evaluation results reveal substantial discrepancies between reanalysis-based metrics and real-world performance, particularly concerning extreme events. By highlighting the limitations of existing benchmarks, this work establishes a more faithful and operationally relevant evaluation paradigm, providing a rigorous foundation for advancing next-generation AI weather forecasting systems. The benchmark implementation is available at: \url{https://github.com/lixruize-del/NWP-Benchmark}.
  
  
\end{abstract}

\RealBenchHighlight{RealBench reveals that strong ERA5 scores do not guarantee operational readiness; station observations and extreme-event metrics expose the largest real-world performance gaps.}

\RealBenchTOC

\section{Introduction}

Accurate weather forecasting is a critical challenge with profound socioeconomic implications, directly impacting sectors such as aviation, maritime navigation, and finance~\cite{dehalwar2016electricity,gultepe2019review,ukhurebor2022precision}. Recent advances in deep learning have enabled AI-based weather forecasting methods~\cite{lam2023graphcast, Price2024Gencast, vamoe, stcast} to achieve performance comparable to traditional numerical weather prediction (NWP) systems, while offering substantially faster inference. Despite this rapid progress, a critical question remains underexplored: \textbf{Do current evaluation protocols faithfully reflect the performance of AI-based models under real-world operational conditions with observational data?}

Currently, the answer is often no. While existing evaluation paradigms \cite{bi2023pangu, rasp2020weatherbench, rasp2024weatherbench2, Pathak2022fourcastnet, nguyen2024stormer, chen2023fuxi} have significantly advanced standardized assessments for AI weather forecasting, they remain limited in their ability to reflect real-world, operational conditions. Several design choices contribute to this discrepancy. 1) Their reliance on relatively early test periods (\textit{e.g.}, 2020) raises concerns regarding potential data leakage and temporal representativeness, as models are not evaluated on the latest atmospheric regimes or emerging distribution shifts. 2) They predominantly adopt ERA5 reanalysis data \cite{hersbach2020era5} as the ground truth rather than real-time operational conditions or real-world station observations. As a model-informed, temporally lagged 4D-Var data assimilation product, ERA5 inherently introduces biases and may underestimate localized extremes, such as temperature peaks and tropical cyclone intensity. 3) Their heavy reliance on global average metrics (\textit{e.g.}, RMSE) emphasizes common large-scale patterns while obscuring performance on rare but high-impact extreme events. Although recent works \cite{jin2024weatherreal, han2024weather5k} have begun incorporating real station observations to better approximate real-world conditions, they rely on a limited number of stations with sparse spatial coverage and evaluate only a small set of models, falling short of a comprehensive and systematic quantification of these biases.


To systematically bridge the gap between benchmark scores and real-world utility, we introduce \textbf{RealBench}, a next-generation benchmark for evaluating AI weather models under realistic operational conditions, which reconstructs the evaluation pipeline by shifting from historical, reanalysis-dependent metrics to operational, observation-grounded assessments. RealBench provides a comprehensive evaluation protocol that isolates models from training data leakage, replaces temporally lagged reference fields with real-time operational analysis and dense global station observations, and introduces targeted protocols specifically formulated to capture high-impact extreme weather events.

We make four core contributions: \textbf{(i) Recent zero-leakage evaluation (2025).} We construct a strictly out-of-distribution test set covering 2025, ensuring no exposure to existing training data. This provides a more up-to-date evaluation that reflects current atmospheric regimes and recent distribution shifts. \textbf{(ii) Evaluation on operational analysis data.} We introduce an ECMWF operational analysis dataset that reflects near real-time forecasting conditions with substantially lower latency than ERA5. Systematically evaluating state-of-the-art data-driven models on this dataset reveals notable discrepancies compared to ERA5-based evaluations, highlighting the necessity of operational data. \textbf{(iii) Evaluation on WEATHER-10K station observations.} We evaluate models using WEATHER-10K, a large-scale global dataset comprising over 10,000 in-situ stations sourced from the Global Historical Climatology Network hourly (GHCNh), complete with rigorous preprocessing (\textit{e.g.}, outlier correction). This enables a more direct assessment against real atmospheric observations with significantly improved spatial coverage. \textbf{(iv) Comprehensive evaluation of extreme events and an open-source framework.} We systematically evaluate extreme, high-impact weather events, including heatwaves, cold surges, and tropical cyclones. For temperature extremes, we adopt a sliding-window approach across a 12-year climatology to define event thresholds (90th/10th percentiles). For tropical cyclones, we evaluate all 2025 events using track, mean sea level pressure (MSL), and wind speed errors. Furthermore, we open-source a user-friendly unified PyTorch inference framework to support seamless evaluation across multiple state-of-the-art AI forecasting models.



Ultimately, RealBench establishes a rigorous and realistic evaluation framework for next-generation AI weather forecasting systems. By explicitly grounding evaluations in real-time constraints, real-world observations, and the precise assessment of high-impact extreme events, RealBench successfully bridges the critical gap between theoretical benchmark performance and practical forecasting utility.

\section{Related Work}

\paragraph{Benchmark Datasets for Weather and Climate Modeling.}
Recent works have developed benchmarks for climate modeling across a range of tasks. WeatherBench~\cite{rasp2020weatherbench} introduced a standardized ERA5-based benchmark for medium-range forecasting and was later extended to probabilistic settings~\cite{garg2022weatherbenchProbability}; WeatherBench2~\cite{rasp2024weatherbench2} further improves resolution, metrics, scalability, and baseline comparisons. Other benchmarks address subseasonal forecasting~\cite{mouatadid2021learned}, precipitation forecasting~\cite{de2021rainbench, sit2021iowarain}, extreme-event analysis~\cite{racah2017extremeweather, rahnemoonfar2021floodnet, requena2021earthnet2021, minixhofer2021droughted, prabhat2020climatenet}, and unified multi-task evaluation~\cite{nguyen2023climatelearn}. Recent datasets such as WeatherReal~\cite{jin2024weatherreal} incorporate in-situ observations, but remain limited in spatial coverage and evaluation scope. Additional benchmarks focus on modeling components and long-term behavior~\cite{cachay2021climart, watson2022climatebench}.

Despite this progress, existing benchmarks exhibit two key limitations: they predominantly rely on reanalysis or simulated data (\textit{e.g.}, ERA5~\cite{hersbach2020era5}, CMIP6~\cite{eyring2016overview}) that fail to capture real-world operational constraints, and their reliance on aggregate metrics often under-characterizes extreme events. To our knowledge, we introduce the first standardized benchmark for AI weather models that systematically unifies medium-range and high-impact forecasting. By evaluating models on real-time, real-world data, our framework enables a more faithful assessment of true operational capabilities.

\paragraph{AI-based Weather Forecasting.} With the rapid rise of deep learning~\cite{he2016deep, chen2023self, ewmoe, oneforecast, chen2024pixel, d3u, continuousensemble, chen2024multi}, data-driven weather modeling has rapidly elevated forecasting capabilities across a diverse range of meteorological tasks. Early milestones primarily concentrated on deterministic global medium-range forecasting \cite{lam2023graphcast, vamoe, sfno, koopmanlab, stcast, bi2023pangu, Pathak2022fourcastnet, fno, emformer}, quickly achieving parity with or surpassing traditional numerical weather prediction (NWP) systems. Building upon these foundations, subsequent research has broadened the operational scope of AI models. This includes the development of generative and ensemble approaches for uncertainty quantification \cite{Price2024Gencast, dyffusion, seeds, aifs_crps, stormcast, difflam, omnicast, dgdm, SDEdit-Weather, SphericalDYffusion, codicast}, as well as the integration of data assimilation directly into end-to-end learning pipelines \cite{Zhang2023Fengwu-4dvar, Chen2023aFengWu-Adas}. Furthermore, the field has expanded to tackle extended forecasting horizons and specialized scenarios, such as high-resolution downscaling \cite{han2024fengwughr}, extreme weather prediction \cite{Chen2024aExtremecast}, and long-term climate simulation \cite{Kochkov2023neuralgcm}. Most recently, the paradigm has shifted toward large-scale foundation and hybrid models \cite{bodnar2025aurora, schmude2024prithvi, Lang2024AIFS, Lang2024bAIFS-CRPS}, which synthesize massive multi-source datasets to provide versatile, general-purpose meteorological forecasting engines.

\begin{figure}[t]
	\centering
    \includegraphics[width=\linewidth]{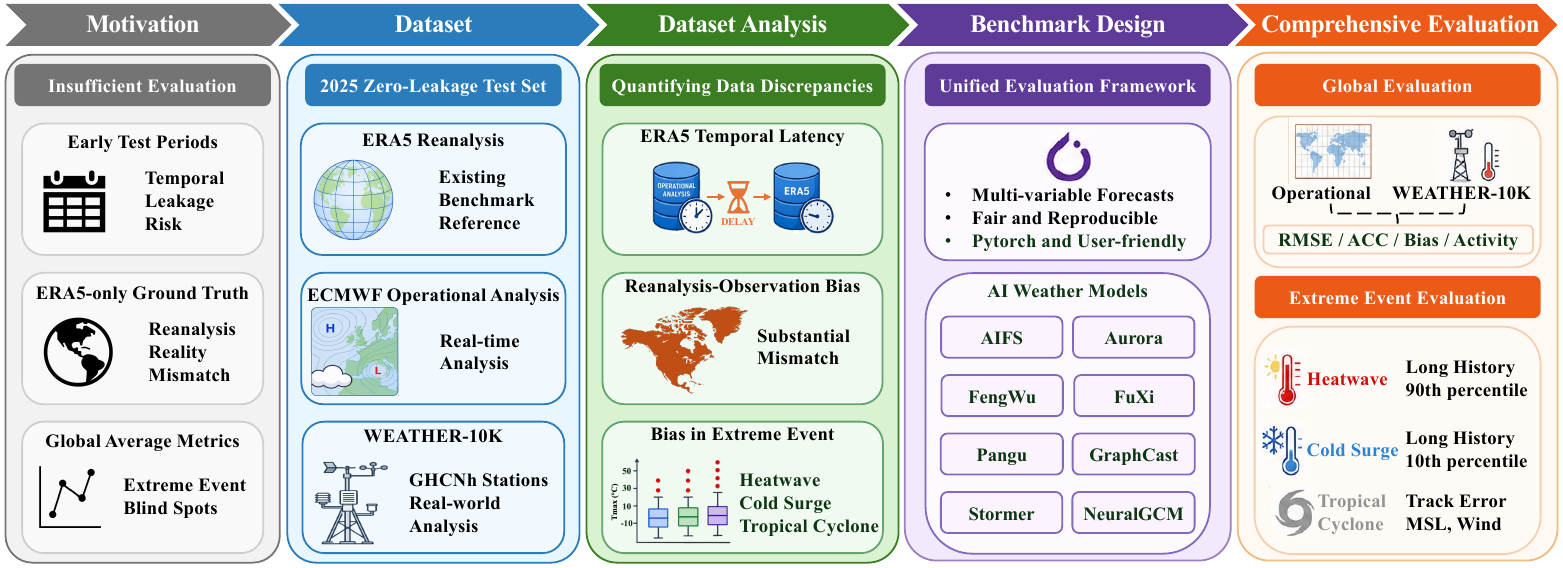}
	\caption{Overview of RealBench, illustrating a real-world and real-time benchmarking pipeline for evaluating AI weather models in operational forecasting settings.}
        \label{fig:framework}
\end{figure}

\section{Dataset}


\paragraph{ERA5 and Operational Analysis Data.}
As shown in Figure~\ref{fig:framework}, we evaluate models on both ERA5 reanalysis~\cite{hersbach2020era5} and ECMWF operational analysis data. ERA5 provides hourly global reanalysis at 0.25$^\circ$ resolution from 1940 to near-present, generated by a 4D-Var assimilation system with 12-hour windows. This extended window improves retrospective consistency but introduces substantial latency and information unavailable under strict real-time forecasting. In contrast, ECMWF operational analyses use a shorter $\pm$3-hour 4D-Var window around initialization, relying on near-real-time observations for low-latency production. Thus, ERA5 serves as a high-quality retrospective reference, whereas operational analysis data better reflect information availability in real-time settings.

\paragraph{WEATHER-10K.} WEATHER-10K is derived from global near-surface in-situ observations provided by the Global Historical Climatology Network hourly dataset (GHCNh). GHCNh is a next-generation hourly dataset consisting of surface weather measurements from fixed, land-based stations, compiled from multiple sources maintained by NOAA, the U.S. Air Force, and other meteorological agencies, and harmonized into a unified format. The resulting dataset contains observations from over 10,000 stations, providing extensive spatial coverage across diverse geographic regions, albeit with substantial geographic imbalance in station density. Further details on station distribution, country-level statistics, and data preprocessing procedures are provided in Appendix~\ref{app:weather10k_details}.

\begin{figure}[t]
	\centering
	\includegraphics[width=\linewidth]{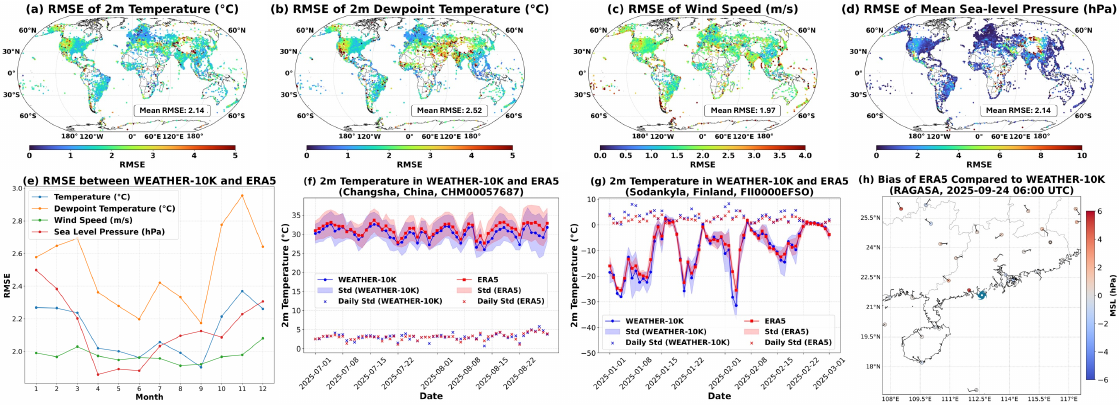}
	\caption{Data analysis and visualizations. (a–d) Global RMSE in 2025 for T2M, D2M, wind speed, and MSL. (e) Monthly RMSE of the four variables in 2025. (f–g) T2M comparisons during heatwave and cold surge periods at Changsha, China, and Sodankylä, Finland. (h) Bias of ERA5 relative to WEATHER-10K in MSL, wind speed, and wind direction for Typhoon Ragasa.}
        \label{fig:data_analysis}
\end{figure}

\paragraph{Dataset Analysis.} As shown in Figure~\ref{fig:data_analysis}(a)--(e), we compute the RMSE between WEATHER-10K and ERA5 in 2025 across four near-surface variables. Overall, T2M and D2M exhibit moderate errors with mean RMSEs of 2.14°C and 2.52°C, respectively, and show larger discrepancies over regions with complex terrain or heterogeneous surface conditions. Wind speed shows comparable mean error magnitude (1.97 m/s) but stronger spatial variability, reflecting the difficulty of resolving near-surface winds affected by local topography, surface roughness, and boundary-layer processes. In contrast, MSL has the lowest overall discrepancy, with a mean RMSE of 2.14 hPa and broadly smoother spatial patterns, consistent with the stronger large-scale constraints on pressure fields. The monthly RMSE curves further indicate that the discrepancies are seasonally dependent, with D2M showing the largest temporal variation among the four variables.

Figure~\ref{fig:data_analysis}(f)--(h) further examines WEATHER-10K and ERA5 under representative extreme-weather conditions. During the summer heatwave period in Changsha, China, and the winter cold surge period in Sodankylä, Finland, the two datasets generally capture similar T2M temporal evolution, while local deviations remain evident at daily time scales. These deviations highlight the challenge of matching station-level temperature extremes, especially when sub-grid surface effects and local meteorological variability become important. For Typhoon Ragasa, the spatial bias map of MSL, wind speed, and wind direction shows structured differences between ERA5 and WEATHER-10K around the storm-affected region, suggesting that discrepancies during tropical cyclones are not purely random but are linked to the representation of storm-scale pressure and wind structures.

Overall, these results reveal variable- and region-dependent gaps between ERA5 and in-situ observations. While ERA5 provides spatially complete and dynamically consistent fields, it can smooth local variability captured by WEATHER-10K, highlighting the need for station-based evaluation.

\section{RealBench: Benchmark Design and Evaluation Settings}
\paragraph{Problem Formulation.} We formulate global medium-range weather forecasting as a spatiotemporal prediction problem over gridded atmospheric states. Let $X_t \in \mathbb{R}^{C \times H \times W}$ denote the atmospheric state at time $t$, where $C$ is the number of atmospheric variables and $H \times W$ is the spatial resolution. Given the current state $X_t$, the objective is to learn a parametric model $F_\theta$ that predicts future states over a horizon $T$, \textit{i.e.}, $\hat{X}_{t+1:t+T} = F_\theta(X_t)$. Forecasting is often performed in an autoregressive manner, where $\hat{X}_{t+k} = F_\theta(\hat{X}_{t+k-1})$ for $k=1,\ldots,T$. The model is trained to minimize the discrepancy between predictions $\hat{X}_{t+k}$ and ground-truth states $X_{t+k}$ over the forecast horizon.

\paragraph{Baselines.} We compare representative state-of-the-art AI-based global weather forecasting models, including AIFS \cite{Lang2024AIFS}, Aurora \cite{bodnar2025aurora}, GraphCast \cite{lam2023graphcast}, Pangu-Weather \cite{bi2023pangu}, FuXi \cite{chen2023fuxi}, FengWu \cite{chen2025operational}, Stormer \cite{nguyen2024stormer}, and NeuralGCM \cite{Kochkov2023neuralgcm}. For all baseline models, inference is conducted using their officially released, open-source checkpoints; the specific model checkpoint versions utilized for both ERA5-based and IFS-based inference in this benchmark are detailed in Table~\ref{tab:baseline_model_versions}. Detailed descriptions of each baseline model architecture are provided in Appendix~\ref{app:baselines}.

\paragraph{Evaluation Setup and Metrics.}
We evaluate model performance on the year 2025 using three datasets: ERA5, operational analysis data, and WEATHER-10K with a 6-hour temporal resolution. For tropical cyclones, we comprehensively evaluate all 72 events occurring from June to December 2025. Forecasts are generated up to a 10-day lead time (\textit{i.e.}, 40 autoregressive steps at 6-hour intervals). To ensure consistent evaluation across models with varying native resolutions, all model outputs are interpolated to a uniform $0.25^\circ$ grid. Following standard practice, we assess performance on eight key atmospheric variables: 2-meter temperature (T2M), 10-meter zonal and meridional winds (U10, V10), mean sea level pressure (MSL), geopotential height at 500 hPa (Z500), temperature at 850 hPa (T850), specific humidity at 700 hPa (Q700), and 850 hPa zonal wind speed (U850). Forecast accuracy is evaluated using four latitude-weighted spatial metrics: weighted root-mean-square error (WRMSE) \cite{han2024fengwughr}, bias, anomaly correlation coefficient (ACC), and activity. Furthermore, we compute the zonal energy spectrum to assess the physical realism of the forecasts across different spatial scales. The full variable list and metric definitions are provided in Appendix~\ref{app:varlist} and Appendix~\ref{app:metrics}. Further implementation details and computational cost comparisons are provided in Appendix~\ref{hardware} and Table~\ref{tab:nwp_efficiency}.


\paragraph{Heatwaves and Cold Surges Evaluation.}
Following prior work on temperature extremes \cite{perkins2013measurement}, 
we evaluate predicted heatwave and cold surge events using a percentile-based event definition and an IoU-based temporal matching strategy. 
Detailed definitions of the extreme-event thresholds, consecutive-day criteria, and event matching procedure are provided in Appendix~\ref{app:metrics}. Based on the matched predicted and ground-truth events, we employ three standard metrics \cite{wilks2011statistical}: Probability of Detection (POD), False Alarm Ratio (FAR), and Critical Success Index (CSI), defined as
\begin{equation}
\mathrm{POD}=\frac{\mathrm{TP}}{\mathrm{TP}+\mathrm{FN}},
\qquad
\mathrm{FAR}=\frac{\mathrm{FP}}{\mathrm{TP}+\mathrm{FP}},
\qquad
\mathrm{CSI}=\frac{\mathrm{TP}}{\mathrm{TP}+\mathrm{FP}+\mathrm{FN}}.
\end{equation}
POD measures the fraction of observed events that were correctly forecast, FAR quantifies the fraction of predicted events that did not occur, and CSI provides a balanced assessment by penalizing both misses and false alarms while ignoring the overwhelmingly large number of true negatives characteristic of rare extreme events.

\begin{figure}[t]
	\centering
	\includegraphics[width=\linewidth]{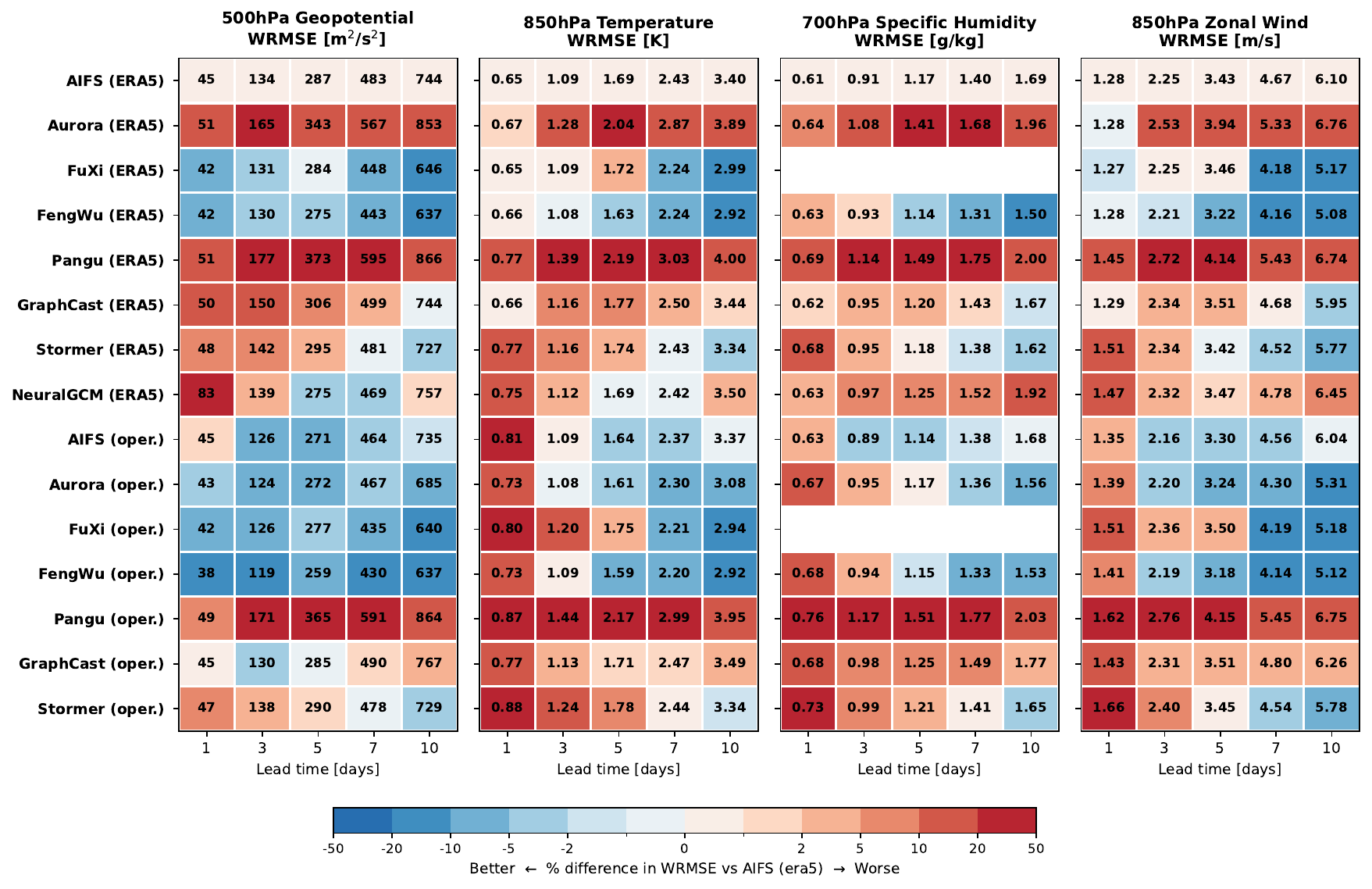}
	\caption{\textbf{Scorecard for upper-level variables for the year 2025.} (oper.) Operational models are evaluated against operational analysis.}
\label{fig:scorecard_upper}
\end{figure}

\paragraph{Tropical Cyclones.} To ensure a fair comparison, the metrics at a specific lead time $l$ are averaged only over the subset of storms, denoted as $\mathcal{S}_l$, that are successfully tracked by all evaluated models. Given the predicted and observed cyclone centers $(\hat{\phi}_{s,l},\hat{\lambda}_{s,l})$ and $(\phi_{s,l},\lambda_{s,l})$ for storm $s$ at lead time $l$, the direct position error (DPE) is measured by the average great-circle distance:
\begin{equation}
\mathrm{DPE}_{l}
=
\frac{1}{|\mathcal{S}_l|}
\sum_{s\in\mathcal{S}_l}
d_{\mathrm{gc}}
\left(
(\hat{\phi}_{s,l},\hat{\lambda}_{s,l}),
(\phi_{s,l},\lambda_{s,l})
\right),
\end{equation}
where $d_{\mathrm{gc}}(\cdot,\cdot)$ denotes the haversine distance on the sphere. 

We additionally evaluate intensity errors using the maximum 10-meter wind speed ($V$) and minimum sea-level pressure ($P^{\min}$). Since data-driven models often suffer from structural under-intensification (\textit{i.e.}, producing smoother fields), we report both the Mean Absolute Error (MAE) and the systematic Bias at each lead time $l$:
\begin{equation}
\mathrm{MAE}^{P}_{l}
=
\frac{1}{|\mathcal{S}_l|}
\sum_{s\in\mathcal{S}_l}
\left|
\hat{P}^{\min}_{s,l}-P^{\min}_{s,l}
\right|,
\qquad
\mathrm{Bias}^{P}_{l}
=
\frac{1}{|\mathcal{S}_l|}
\sum_{s\in\mathcal{S}_l}
\left(
\hat{P}^{\min}_{s,l}-P^{\min}_{s,l}
\right).
\end{equation}
The wind speed errors ($\mathrm{MAE}^{V}_{l}$ and $\mathrm{Bias}^{V}_{l}$) are formulated analogously.

\section{Experimental Results}




\subsection{Comparison of Medium-range Weather Forecasts}

\begin{figure}[t]
	\centering
	\includegraphics[width=\linewidth]{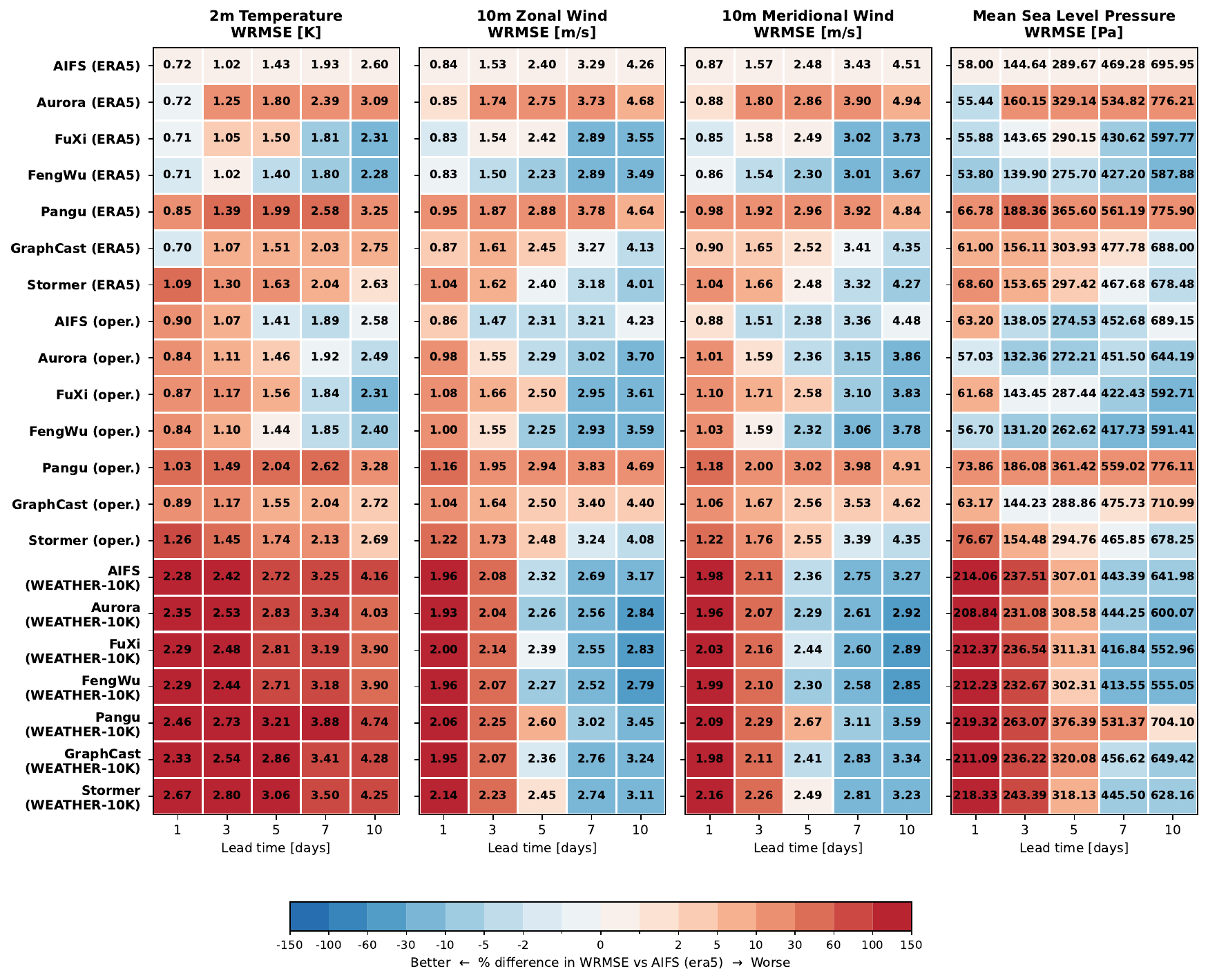}
	\caption{\textbf{Scorecard for surface-level variables for the year 2025.} (oper.) Operational models are evaluated against operational analysis. (WEATHER-10K) WRMSE computed between station observations and operational forecast outputs interpolated to station locations.}
        \label{fig:scorecard_surface}
\end{figure}

\begin{figure}[t]
	\centering
	\includegraphics[width=\linewidth]{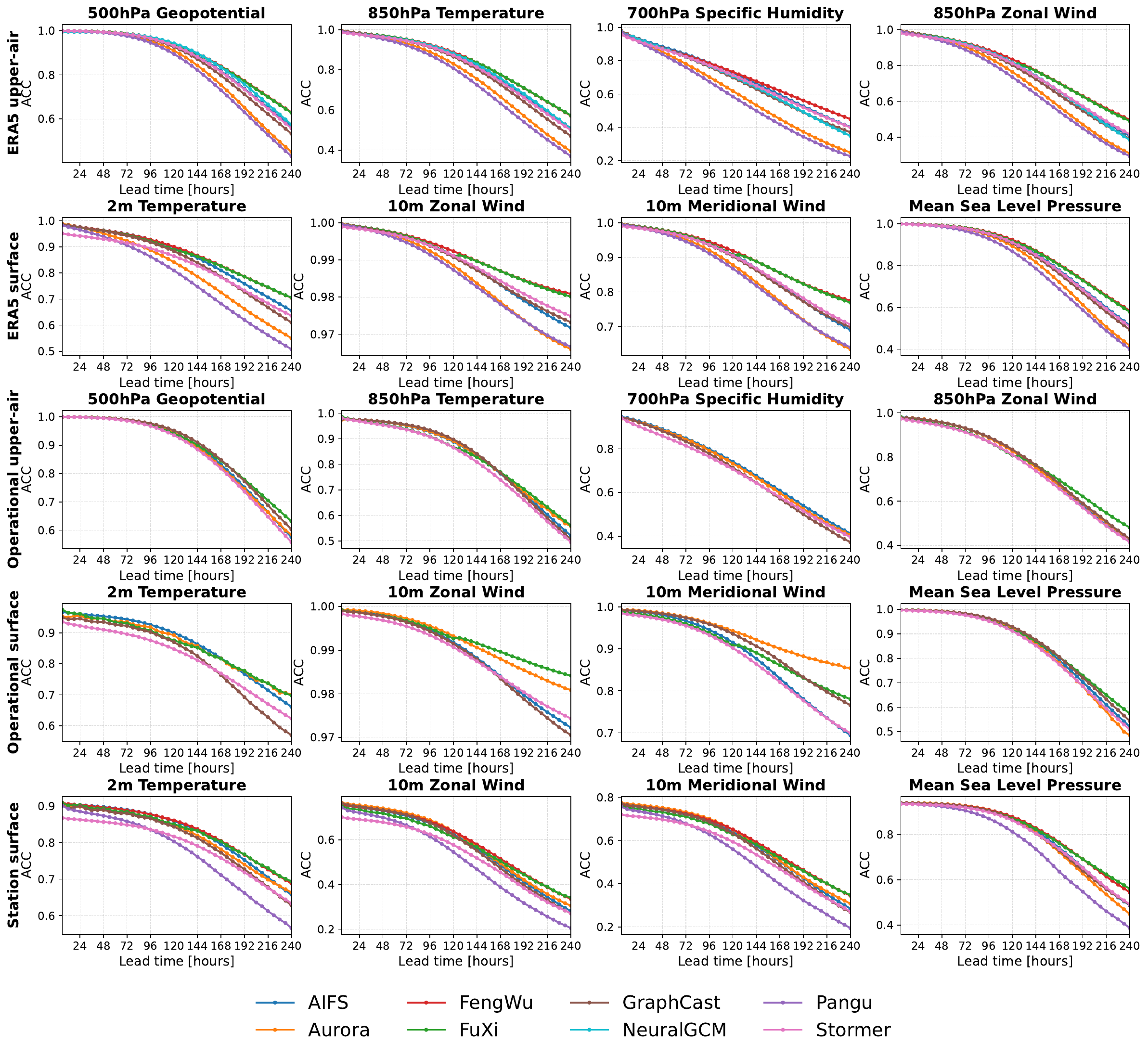}
	\caption{ACC $\uparrow$ comparison of eight baseline models across ERA5, operational analysis data, and WEATHER-10K station evaluations in global weather forecasting.}
\label{fig:acc_curves}
\end{figure}

\paragraph{Upper-level Evaluation.} As shown in Figure~\ref{fig:scorecard_upper}, FengWu and FuXi achieve the strongest overall performance across most upper-level variables and lead times, especially at medium-to-long ranges. For Z500, FengWu (oper.) obtains the lowest RMSE from 1 to 10 days, with values of 38, 119, 259, 430, and 637 $\text{m}^2/\text{s}^2$, while FuXi (oper.) remains highly competitive, reaching 435 and 640 $\text{m}^2/\text{s}^2$ at 7 and 10 days, respectively. For T850, FengWu (oper.) also achieves the best or near-best results, with RMSEs of 2.20 K and 2.92 K at 7 and 10 days. Similar trends are observed for Q700 and U850, where FengWu maintains strong long-range accuracy. Overall, the upper-level results indicate that operational-analysis evaluation preserves the relative ranking of leading models, with FengWu and FuXi consistently providing the most accurate forecasts among the compared systems.


\paragraph{Surface Evaluation.} As shown in Figure~\ref{fig:scorecard_surface}, the results include ERA5, operational-analysis, and WEATHER-10K station-based evaluations, where `oper.' denotes operational analysis data. On ERA5 and operational-analysis data, FengWu and FuXi remain the strongest models across most variables, especially at longer lead times. For example, FengWu (ERA5) achieves 2.28 K for 2-meter temperature (T2M), 3.49 m/s for 10-meter zonal wind (U10), 3.67 m/s for 10-meter meridional wind (V10), and 587.88 Pa for mean sea-level pressure (MSL) at 10 days. Under the operational setting, FengWu obtains 2.40 K, 3.59 m/s, 3.78 m/s, and 591.41 Pa at 10 days for the same variables, while FuXi is particularly competitive for T2M and MSL\@. On WEATHER-10K station evaluation, the RMSE is larger, reflecting the increased difficulty of pointwise station verification. This discrepancy is expected, since a natural mismatch already exists between analysis data and station observations (as shown in Figure~\ref{fig:data_analysis}). Consequently, AI models trained with ERA5 as the ground truth inevitably inherit such discrepancies when evaluated against real station measurements. This observation further indicates that forecasting on irregularly distributed stations is inherently challenging, and highlights the necessity of station-based evaluation under real-world conditions.


\paragraph{Cross-Dataset ACC Comparison.} As shown in Figure~\ref{fig:acc_curves}, we compare AI-based models using ACC across ERA5, operational analysis data, and WEATHER-10K. ACC decreases with increasing lead time, and performance gaps become more pronounced at longer horizons. Compared with gridded ERA5 and IFS evaluations, station-based surface evaluation yields lower ACC, especially for near-surface winds, indicating the challenge of forecasting on irregularly distributed stations and the necessity of real-world station-based evaluation. More results are presented in the Appendix~\ref{app:more_result}.


\subsection{Extreme Events Assessment}
\begin{table}[t]
\centering
\footnotesize
\setlength{\tabcolsep}{2pt}
\renewcommand{\arraystretch}{1.0}
\begin{tabular*}{0.96\textwidth}{@{\extracolsep{\fill}}llcccc@{\hspace{1.2em}}cccc@{}}
\toprule
\multirow{2}{*}{\textbf{Model}} 
& \multirow{2}{*}{\textbf{Metric}} 
& \multicolumn{4}{c}{\textbf{Heatwave (2025 H2)}} 
& \multicolumn{4}{c}{\textbf{Cold Surge (2025 H2)}} \\
\cmidrule(lr){3-6} \cmidrule(lr){7-10}
& & \textbf{Day 1} & \textbf{Day 3} & \textbf{Day 7} & \textbf{Day 10} 
  & \textbf{Day 1} & \textbf{Day 3} & \textbf{Day 7} & \textbf{Day 10} \\
\midrule

\multirow{3}{*}{AIFS} 
& CSI $\uparrow$ & \textbf{0.513} & \textbf{0.535} & \textbf{0.353} & \textbf{0.215} & \textbf{0.370} & \textbf{0.413} & \textbf{0.234} & \textbf{0.131} \\
& POD $\uparrow$ & \textbf{0.646} & \textbf{0.689} & \textbf{0.533} & \textbf{0.399} & \textbf{0.443} & 0.526 & 0.348 & \textbf{0.249} \\
& FAR $\downarrow$ & \textbf{0.291} & \textbf{0.311} & 0.489 & 0.673 & \textbf{0.294} & \textbf{0.347} & 0.585 & 0.774 \\
\midrule

\multirow{3}{*}{FengWu} 
& CSI $\uparrow$ & 0.486 & 0.486 & 0.279 & 0.136 & 0.358 & 0.390 & 0.196 & 0.060 \\
& POD $\uparrow$ & 0.615 & 0.628 & 0.368 & 0.189 & 0.432 & 0.522 & 0.260 & 0.075 \\
& FAR $\downarrow$ & 0.301 & 0.327 & \textbf{0.447} & \textbf{0.620} & 0.312 & 0.394 & \textbf{0.548} & \textbf{0.723} \\
\midrule

\multirow{3}{*}{FuXi} 
& CSI $\uparrow$ & 0.470 & 0.423 & 0.283 & 0.168 & 0.344 & 0.372 & 0.201 & 0.064 \\
& POD $\uparrow$ & 0.632 & 0.685 & 0.472 & 0.281 & 0.433 & 0.544 & 0.295 & 0.092 \\
& FAR $\downarrow$ & 0.355 & 0.483 & 0.573 & 0.672 & 0.356 & 0.459 & 0.609 & 0.795 \\
\midrule

\multirow{3}{*}{Pangu} 
& CSI $\uparrow$ & 0.458 & 0.381 & 0.165 & 0.086 & 0.342 & 0.290 & 0.075 & 0.028 \\
& POD $\uparrow$ & 0.628 & 0.598 & 0.296 & 0.172 & 0.410 & \textbf{0.599} & \textbf{0.414} & 0.244 \\
& FAR $\downarrow$ & 0.373 & 0.479 & 0.667 & 0.782 & 0.313 & 0.633 & 0.909 & 0.965 \\
\midrule

\multirow{3}{*}{Aurora} 
& CSI $\uparrow$ & 0.450 & 0.443 & 0.252 & 0.132 & 0.331 & 0.348 & 0.132 & 0.024 \\
& POD $\uparrow$ & 0.586 & 0.606 & 0.347 & 0.199 & 0.417 & 0.513 & 0.204 & 0.039 \\
& FAR $\downarrow$ & 0.332 & 0.379 & 0.483 & 0.668 & 0.368 & 0.474 & 0.715 & 0.922 \\
\midrule

\multirow{3}{*}{GraphCast} 
& CSI $\uparrow$ & 0.437 & 0.392 & 0.259 & 0.160 & 0.316 & 0.328 & 0.200 & 0.085 \\
& POD $\uparrow$ & 0.588 & 0.595 & 0.452 & 0.326 & 0.397 & 0.488 & 0.350 & 0.204 \\
& FAR $\downarrow$ & 0.364 & 0.448 & 0.592 & 0.732 & 0.383 & 0.501 & 0.681 & 0.865 \\
\midrule

\multirow{3}{*}{Stormer} 
& CSI $\uparrow$ & 0.426 & 0.432 & 0.275 & 0.169 & 0.305 & 0.332 & 0.183 & 0.087 \\
& POD $\uparrow$ & 0.536 & 0.574 & 0.403 & 0.277 & 0.388 & 0.456 & 0.291 & 0.174 \\
& FAR $\downarrow$ & 0.313 & 0.364 & 0.510 & 0.661 & 0.381 & 0.437 & 0.655 & 0.838 \\

\bottomrule
\end{tabular*}
\caption{Detailed verification metrics (CSI, POD, FAR) across $1$ to $10$ lead days with multiple models.}
\label{tab:detailed_verification_metrics}
\end{table}

\paragraph{Heatwaves.} 
Table~\ref{tab:detailed_verification_metrics} details the performance on global heatwave events during the second half of 2025. Overall, forecasting skill declines as lead time increases, with AIFS consistently performing best. Specifically, we observe the following key phenomena: \textbf{(i) Precision-Recall Trade-offs.} The detailed metrics reveal varying characteristics among the architectures. Most models exhibit a conservative tendency, maintaining relatively high precision (low False Alarm Ratio, FAR) but suffering from lower recall (Probability of Detection, POD) at early lead times. FengWu exemplifies this robustness, maintaining the lowest FAR globally at extended lead times (\textit{e.g.}, 0.447 at Day 7). This suggests that models hesitate to predict extreme heat objects unless the large-scale circulation signal is overwhelmingly strong. \textbf{(ii) Detail Loss at Long Lead Times.} Beyond Day 7, forecasting skill for extreme heat declines significantly. By Day 10, most models drop below a CSI of 0.17 (\textit{e.g.}, Pangu-Weather drops to 0.086). This performance degradation may be attributed to the tendency of RMSE optimization to smooth spatial gradients, which underestimates heatwave intensity at extended lead times. This smoothing effect is empirically demonstrated by the systematic negative biases in the case studies (see Appendix Figure~\ref{fig:heatwave}).

\paragraph{Cold Surges.} 
A direct comparison in Table~\ref{tab:detailed_verification_metrics} reveals that the performance degradation observed in heatwaves is also present in cold surges, but the overall performance is worse. For instance, the Day 1 CSI for the best-performing model (AIFS) drops from 0.513 (heatwaves) to 0.370 (cold surges). At extended lead times (Day 7 and Day 10), the FAR for cold surges increases rapidly across multiple models. Interestingly, while Pangu-Weather achieves the highest POD at Day 3 (0.599) and Day 7 (0.414) for cold surges, it simultaneously produces a FAR of 0.633 and 0.909, respectively. It captures cold events through over-prediction, resulting in a large number of false positives among its predicted cold objects.

This asymmetric performance may stem from an imbalanced target distribution in the training data. Under the recent global warming trend, extreme cold anomalies relative to the 1979--1990 climatological baseline are significantly rarer in the recent training data than extreme heatwaves. Consequently, models are generally less capable of forecasting cold surges. When predicting these events, models tend to overly smooth the outputs. This can either cause forecasts to miss the strict P10 threshold (leading to low POD), or generate temporally and spatially smoothed events that fail to overlap with the ground truth under the temporal IoU metric (leading to high FAR). As shown in Appendix Figure~\ref{fig:coldwave}, this smoothing causes an overall positive bias during cold surges (\textit{i.e.}, predicting temperatures too warm for cold extremes).

\paragraph{Tropical Cyclones.}
\begin{table}[t]
\centering
\footnotesize
\setlength{\tabcolsep}{2pt}
\renewcommand{\arraystretch}{0.95}
\begin{tabular*}{0.95\textwidth}{@{\extracolsep{\fill}}llccc@{\hspace{2pt}}ccc@{}}
\toprule
\multirow{2}{*}{\textbf{Model}} 
& \multirow{2}{*}{\textbf{Metric}} 
& \multicolumn{3}{c}{\textbf{ERA5 Initialized}} 
& \multicolumn{3}{c}{\textbf{IFS Initialized}} \\
\cmidrule(lr){3-5} \cmidrule(lr){6-8}
& & \textbf{Day 1} & \textbf{Day 3} & \textbf{Day 5}
  & \textbf{Day 1} & \textbf{Day 3} & \textbf{Day 5} \\
\midrule

\multirow{4}{*}{AIFS} 
& Track DPE (km) $\downarrow$ & \textbf{54.7} & 154.1 & 428.3 & 52.7 & 138.2 & 366.3 \\
& MSLP MAE (hPa) $\downarrow$ & \textbf{11.5} & \textbf{16.8} & 18.2 & \textbf{10.7} & \textbf{16.1} & \textbf{16.2} \\
& MSLP Bias (hPa) & \textbf{+11.1} & +16.2 & +16.8 & \textbf{+10.1} & \textbf{+15.4} & \textbf{+14.3} \\
& Wind MAE (m/s) $\downarrow$ & 14.3 & 17.8 & 17.6 & 13.6 & \textbf{17.4} & \textbf{15.9} \\
\midrule

\multirow{4}{*}{Aurora} 
& Track DPE (km) $\downarrow$ & 55.7 & \textbf{148.7} & 427.6 & \textbf{51.6} & \textbf{133.9} & 338.1 \\
& MSLP MAE (hPa) $\downarrow$ & 12.3 & 17.3 & \textbf{17.8} & 11.4 & 20.1 & 29.1 \\
& MSLP Bias (hPa) & +11.9 & \textbf{+15.2} & \textbf{+15.1} & +10.8 & +20.0 & +29.1 \\
& Wind MAE (m/s) $\downarrow$ & \textbf{13.9} & \textbf{16.4} & \textbf{15.2} & \textbf{13.5} & 19.5 & 22.5 \\
\midrule

\multirow{4}{*}{FengWu} 
& Track DPE (km) $\downarrow$ & 58.3 & 163.9 & 409.4 & 56.2 & 140.6 & 371.0 \\
& MSLP MAE (hPa) $\downarrow$ & 14.4 & 24.2 & 39.5 & 13.1 & 21.9 & 29.9 \\
& MSLP Bias (hPa) & +14.3 & +24.1 & +39.5 & +12.8 & +21.5 & +29.8 \\
& Wind MAE (m/s) $\downarrow$ & 15.9 & 22.6 & 29.5 & 14.9 & 21.0 & 24.7 \\
\midrule

\multirow{4}{*}{FuXi} 
& Track DPE (km) $\downarrow$ & 57.3 & 152.4 & 360.2 & 78.1 & 181.7 & 331.0 \\
& MSLP MAE (hPa) $\downarrow$ & 13.8 & 20.0 & 22.8 & 13.8 & 19.7 & 18.2 \\
& MSLP Bias (hPa) & +13.6 & +19.5 & +22.6 & +13.5 & +19.2 & +17.9 \\
& Wind MAE (m/s) $\downarrow$ & 15.3 & 19.2 & 19.0 & 14.6 & 18.8 & 16.8 \\
\midrule

\multirow{4}{*}{GraphCast} 
& Track DPE (km) $\downarrow$ & 59.1 & 149.1 & 320.3 & 52.9 & 135.4 & 371.0 \\
& MSLP MAE (hPa) $\downarrow$ & 13.7 & 21.1 & 26.1 & 11.1 & 18.4 & 17.8 \\
& MSLP Bias (hPa) & +13.5 & +20.7 & +26.1 & +10.3 & +17.3 & +16.2 \\
& Wind MAE (m/s) $\downarrow$ & 15.9 & 21.0 & 22.2 & 13.9 & 18.5 & 16.9 \\
\midrule

\multirow{4}{*}{Pangu} 
& Track DPE (km) $\downarrow$ & 64.5 & 185.9 & 444.7 & 66.0 & 189.2 & 428.8 \\
& MSLP MAE (hPa) $\downarrow$ & 14.5 & 21.3 & 25.9 & 14.3 & 20.3 & 20.3 \\
& MSLP Bias (hPa) & +14.3 & +20.7 & +25.7 & +14.1 & +19.8 & +19.9 \\
& Wind MAE (m/s) $\downarrow$ & 15.2 & 19.2 & 20.1 & 14.9 & 18.7 & 17.1 \\
\midrule

\multirow{4}{*}{Stormer} 
& Track DPE (km) $\downarrow$ & 69.5 & 155.2 & \textbf{320.0} & 68.9 & 151.3 & \textbf{304.9} \\
& MSLP MAE (hPa) $\downarrow$ & 15.8 & 23.3 & 29.0 & 15.4 & 21.9 & 22.8 \\
& MSLP Bias (hPa) & +15.7 & +23.2 & +29.0 & +15.3 & +21.8 & +22.7 \\
& Wind MAE (m/s) $\downarrow$ & 17.7 & 22.2 & 23.4 & 17.2 & 21.6 & 20.8 \\

\bottomrule
\end{tabular*}
\caption{Typhoon track and intensity verification over a homogeneous sample set for ERA5- and IFS-initialized forecasts at 1-, 3-, and 5-day lead times.}
\label{tab:tc_metrics}
\end{table}

Table~\ref{tab:tc_metrics} presents the homogeneous sample evaluation of tropical cyclone track and intensity forecasts during the second half of 2025. In terms of track prediction (Track DPE), all models demonstrate high accuracy at early lead times (Day 1), with errors ranging between 50 and 70 km. As the forecast lead time extends to Day 5, model performance diverges; notably, GraphCast and Stormer maintain relatively low track errors of approximately 300--320 km. Furthermore, comparing the results initialized with ERA5 and IFS reveals that models fine-tuned on operational IFS data (\textit{e.g.}, Aurora and AIFS) exhibit significant reductions in track errors when evaluated with IFS initializations. For instance, Aurora's Day-5 track error drops from 427.6 km (ERA5) to 338.1 km (IFS). Conversely, models without specific operational adaptation (\textit{e.g.}, Pangu-Weather) perform similarly across both initialization datasets. This directly demonstrates the necessity of conducting benchmark evaluations under realistic operational conditions.

Compared to track prediction, current data-driven models generally exhibit notable shortcomings in forecasting cyclone intensity. Across all lead times, the evaluated models produce substantial positive biases in minimum sea level pressure (MSLP biases ranging from +10.1 to +39.5 hPa), along with considerable wind speed errors. This indicates a systematic underestimation of the true cyclone intensity. Notably, the magnitude of this intensity degradation varies among models, largely constrained by their different resolutions. For example, Stormer, which operates at a coarser native resolution of $1.4^\circ$, exhibits the largest MSLP bias. These results suggest that while current AI models can accurately capture tropical cyclone trajectories, precisely representing and maintaining their extreme intensity remains a core challenge for future development.



\section{Discussion}

The primary objective of this work is to evaluate the forecasting performance of data-driven global weather models under operational constraints and against in-situ station observations~\cite{rasp2024weatherbench2, jin2024weatherreal}. Reanalysis datasets such as ERA5, although widely used for benchmarking, are generated through delayed data assimilation and do not fully reflect the constraints of real-time operational forecasting~\cite{hersbach2020era5}. To address this discrepancy, we present \textit{RealBench}, an evaluation framework that integrates near-real-time ECMWF operational analysis and a global ground-based observation dataset (\textit{WEATHER-10K}). The evaluation reveals substantial discrepancies between reanalysis-based metrics and operational performance, particularly during extreme events, thereby bridging the gap between theoretical benchmark performance and practical forecasting utility~\cite{rasp2024weatherbench2, jin2024weatherreal}.

Our results indicate a systematic performance degradation when transitioning from ERA5-based verification to operational analyses. Reanalysis products utilize a temporally lagged, retrospective data assimilation process (such as a 12-hour 4D-Var window) that optimizes global dynamical consistency by incorporating retrospective observations~\cite{hersbach2020era5}. Data-driven models optimized primarily on these highly consistent, smoothed reanalysis fields exhibit a degree of overfitting to these idealized gridded structures. Consequently, when evaluated against near-real-time ECMWF operational analysis, which operates under strict latency constraints and a shorter assimilation window, or against high-frequency physical ground observations, these models exhibit a drop in accuracy. This indicates that theoretical excellence on retrospectively consistent reanalysis fields does not consistently translate to real-world forecasting utility under operational constraints.

These performance discrepancies are further pronounced during high-impact extreme weather events. In our evaluation of global heatwaves, several models show increasingly negative $T_{\text{max}}$ biases at longer lead times during the strongest events, indicating a reduction in peak intensity fidelity as lead time increases (as shown in Figure~\ref{fig:heatwave}). Conversely, for global cold surges, the model spread increases with lead time, and a positive $T_{\text{min}}$ bias becomes more common at longer leads, corresponding to weaker simulated cold intensity relative to ERA5 (as shown in Figure~\ref{fig:coldwave}). This systematic underestimation of extreme peaks is consistent with the spatial smoothing effect induced by optimizing global spatial fields using mean-squared error (MSE) or weighted grid-cell loss functions. Such mathematical optimizations naturally suppress localized peak gradients and high-frequency temporal variances, leading to a general underestimation of extreme value magnitudes.

Several distinct limitations of the \textit{RealBench} framework must be outlined. First, the geographic density of the \textit{WEATHER-10K} ground station network is highly imbalanced (as shown in Figure~\ref{fig:weather10k_station_distribution}), with dense coverage in North America and Europe but sparse distribution over oceans and high-latitude regions, which constrains the global representativeness of the pointwise metrics. Second, there is a fundamental spatial scale mismatch between the $0.25^\circ$ gridded model outputs and pointwise physical measurements, which are strongly influenced by sub-grid-scale local topography and surface roughness. Third, to maintain computational efficiency, the temperature extreme thresholds were computed using a 12-year moving climatological window rather than a standard 30-year meteorological baseline, potentially introducing localized regional biases~\cite{perkins2013measurement}.

To address these limitations, future data-driven weather forecasting frameworks should explore hybrid grid-and-point evaluation metrics and local downscaling post-processing. Specifically, integrating statistical lapse-rate temperature adjustments can partially reconcile the scale discrepancy between grid cells and local stations. Furthermore, model training could incorporate hybrid loss functions that evaluate both grid-scale atmospheric dynamics and point-scale station variations. Future expansions of \textit{RealBench} should focus on incorporating regional high-resolution datasets and probabilistic ensemble systems for operational uncertainty quantification.

\section{Conclusion}

In this work, we introduced RealBench, a unified benchmark that evaluates data-driven medium-range weather forecasting under zero-leakage 2025 testing, operational analysis, station observations, and extreme-event scenarios. First, RealBench replaces ERA5-only evaluation with ECMWF operational analysis and WEATHER-10K, enabling more realistic verification. Second, RealBench extends evaluation from aggregate forecasting accuracy to high-impact events, including heatwaves, cold surges, and tropical cyclones, using event-specific metrics that reflect practical forecasting priorities. Experiments across representative models reveal substantial discrepancies between reanalysis-based and observation-grounded evaluations, especially for near-surface variables and extremes, highlighting RealBench as a rigorous foundation for advancing operationally reliable forecasts.

\bibliographystyle{ieee_fullname}
\bibliography{reference}

\newpage
\appendix

\section{Limitations}

Although RealBench provides a more realistic benchmark for AI weather forecasting, several limitations remain. First, WEATHER-10K is based on globally distributed in-situ stations, but the station network is geographically imbalanced. Stations are denser in regions with well-developed meteorological infrastructure, such as North America, Europe, and parts of Asia, while coverage is sparser over oceans, deserts, high-latitude regions, and parts of Africa and South America. As a result, station-based evaluation may provide more reliable estimates in observation-dense regions than in sparsely observed regions.

Second, station observations and gridded analysis fields are not perfectly comparable because they represent different spatial scales. ERA5 and operational analysis data provide spatially complete and dynamically consistent atmospheric fields, whereas in-situ stations capture local surface conditions affected by terrain, land use, surface roughness, and boundary-layer processes. This scale mismatch is particularly important for near-surface variables and extreme events, and may lead to discrepancies that cannot be attributed solely to model forecast error.

Third, the evaluation of heatwaves and cold surges depends on the chosen event definition and matching protocol. Although the percentile-based threshold, consecutive-day criterion, and temporal-IoU matching strategy are designed to capture high-impact temperature extremes, the resulting scores may vary with the climatological baseline, threshold percentile, minimum event duration, and IoU threshold. Therefore, these metrics should be interpreted as one operationally motivated view of extreme-event skill rather than a unique definition of event predictability.

Finally, our benchmark evaluates existing data-driven forecasting systems under a unified protocol, but does not remove all differences among their native resolutions, input variables, training data, or operational adaptation. Tropical cyclone intensity prediction remains especially challenging: current AI models often capture storm tracks more reliably than storm intensity, and their intensity errors are affected by spatial resolution and smoothing of extreme pressure and wind structures. Future work should extend RealBench to additional years, more observation sources, probabilistic forecasts, and higher-resolution regional verification.

\section{Broader Impacts}

RealBench has several positive societal implications. Accurate and reliable weather forecasting is important for many sectors, including aviation, maritime navigation, finance, energy, agriculture, disaster preparedness, and public safety. By evaluating AI weather models under operational conditions and against real-world station observations, RealBench can help researchers and practitioners better understand whether benchmark performance translates into real-world forecasting utility. Its explicit focus on high-impact events, including heatwaves, cold surges, and tropical cyclones, may support the development of more reliable forecasting systems for weather-risk mitigation and early warning.

At the same time, the benchmark should not be interpreted as a guarantee that any evaluated model is safe for direct operational deployment. Weather forecasts can influence high-stakes decisions, and incorrect predictions may lead to economic loss, inadequate preparation, or misplaced public trust. This is particularly important for extreme events, where our results show that current AI models may smooth temperature extremes, underperform on cold surges, or underestimate tropical cyclone intensity. Therefore, AI-based forecasts should be used with appropriate expert oversight, uncertainty communication, and comparison against established operational forecasting systems.

RealBench also highlights potential regional inequities in weather-model evaluation. Because in-situ station coverage is geographically imbalanced, station-based verification is more representative in observation-dense regions than in sparsely observed areas. Future benchmark development should incorporate additional observation sources and regional datasets to better assess forecasting performance in under-observed regions.

\section{Details of WEATHER-10K}
\label{app:weather10k_details}

\subsection{Station Distribution and Country-level Statistics}

\begin{figure}[t]
	\centering
	\includegraphics[width=\linewidth]{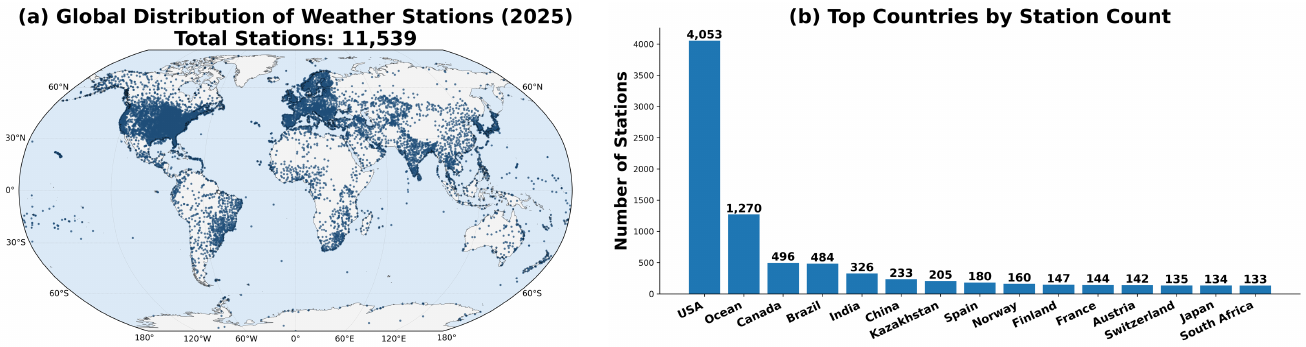}
	\caption{Station distribution of WEATHER-10K. (a) Global locations of all 11,539 stations, showing dense coverage over North America, Europe, and parts of Asia, and relatively sparse coverage over oceans, deserts, high-latitude regions, and parts of Africa and South America. (b) Top countries or regions by station count.}
\label{fig:weather10k_station_distribution}
\end{figure}

Figure~\ref{fig:weather10k_station_distribution} summarizes the spatial distribution and country-level statistics of WEATHER-10K stations. WEATHER-10K contains 11,539 globally distributed surface weather stations, providing broad coverage over North America, Europe, East and South Asia, parts of South America, Africa, and Oceania. The station network is not spatially uniform: observations are denser in regions with well-developed meteorological infrastructure, especially the United States and Europe, while coverage is relatively sparse over deserts, high-latitude regions, oceans, and parts of Africa and South America. At the country level, the United States contributes the largest number of stations, followed by ocean-based stations, Canada, Brazil, India, China, Kazakhstan, Spain, and several European and East Asian countries. This distribution reflects the heterogeneous availability of global in-situ observations and motivates evaluating forecasting models under realistic, irregularly distributed station coverage.

\subsection{Data Preprocessing}

We convert raw station observations into 6-hourly samples. For each target time $t$, we collect observations within a $\pm 15$ minute window and average all valid records in the window:
\begin{equation}
    y_{i,t}^{v}
    =
    \frac{1}{|\mathcal{O}_{i,t}^{v}|}
    \sum_{\tau \in \mathcal{O}_{i,t}^{v}} x_{i,\tau}^{v},
    \quad
    \mathcal{O}_{i,t}^{v}=\{\tau: |\tau-t|\leq 15~\mathrm{min}\},
\end{equation}
where $x_{i,\tau}^{v}$ denotes the raw observation of variable $v$ at station $i$ and time $\tau$. 
Stations without valid observations in the window are excluded for that timestamp.

\subsection{Quality Control with ERA5 Reference Fields}

We further apply an ERA5-based quality-control step to remove extreme station outliers. For each 6-hourly timestamp, ERA5 single-level fields on a $0.25^\circ$ grid are bilinearly interpolated to station locations. Temperature and dew-point temperature are converted from Kelvin to degrees Celsius, pressure variables from Pa to hPa, and ERA5 wind speed is computed from 10-meter wind components as $\sqrt{u_{10}^{2}+v_{10}^{2}}$. We denote the processed station observation by $y_{i,t}^{v}$ and the corresponding station-interpolated ERA5 reference by $\hat{y}_{i,t}^{v}$. For temperature, dew-point temperature, station-level pressure, sea-level pressure, and wind speed, we flag an observation as anomalous if
\begin{equation}
    \frac{y_{i,t}^{v}}{\hat{y}_{i,t}^{v}} > r_v ,
\end{equation}
where $r_v$ is a variable-specific threshold. 
We use $r_v=6$ for temperature and dew-point temperature, $r_v=9000$ for sea-level pressure, and $r_v=7$ for wind speed. 
Flagged values are replaced by the corresponding ERA5-interpolated values, while all other station observations are kept unchanged. This procedure removes severe physically implausible values while preserving the original in-situ observations whenever they pass the quality-control criterion.

\begin{table}[t]
\centering
\small

\resizebox{\textwidth}{!}{
\begin{tabular}{p{2.2cm} p{9.8cm} p{4.6cm}}
\toprule
\textbf{Model} & \textbf{Input variables} & \textbf{Vertical levels} \\
\midrule

AIFS &
Pressure-level variables: geopotential (Z), specific humidity (Q), temperature (T), zonal wind (U), meridional wind (V), vertical velocity (W).

Single-level variables: 2m temperature (T2M), 10m zonal wind (U10), 10m meridional wind (V10), mean sea-level pressure (MSL), surface pressure (SP), total column water (TCW), skin temperature (SKT), 2m dewpoint temperature (D2M), total precipitation (TP), convective precipitation (CP), land-sea mask (LSM), standard deviation of sub-grid orography (SDOR), slope of sub-grid orography (SLOR), total cloud cover (TCC), geopotential at surface (Z), soil temperature level 1 (STL1), soil temperature level 2 (STL2), volumetric soil water layer 1 (SWVL1), volumetric soil water layer 2 (SWVL2). & Pressure levels:
50, 100, 150, 200, 250, 300, 400, 500, 600, 700, 850, 925, 1000 hPa.
 \\

\midrule

Aurora &
Pressure-level variables: geopotential (Z), specific humidity (Q), temperature (T), zonal wind (U), meridional wind (V).

Surface variables: mean sea-level pressure (MSL), 10m zonal wind (U10), 10m meridional wind (V10), 2m temperature (T2M). &
Pressure levels:
50, 100, 150, 200, 250, 300, 400, 500, 600, 700, 850, 925, 1000 hPa. \\

\midrule

GraphCast &
Pressure-level variables: geopotential (Z), specific humidity (Q), temperature (T), zonal wind (U), meridional wind (V), vertical velocity (W).

Surface variables: mean sea-level pressure (MSL), 10m zonal wind (U10), 10m meridional wind (V10), 2m temperature (T2M), 6-hour accumulated total precipitation (TP6H). &
Pressure levels:
1, 2, 3, 5, 7, 10, 20, 30, 50, 70, 100, 125, 150, 175, 200, 225, 250, 300, 350, 400, 450, 500, 550, 600, 650, 700, 750, 775, 800, 825, 850, 875, 900, 925, 950, 975, 1000 hPa. \\

\midrule

Pangu-Weather &
Pressure-level variables: geopotential (Z), specific humidity (Q), temperature (T), zonal wind (U), meridional wind (V).

Surface variables: mean sea-level pressure (MSL), 10m zonal wind (U10), 10m meridional wind (V10), 2m temperature (T2M). &
Pressure levels:
50, 100, 150, 200, 250, 300, 400, 500, 600, 700, 850, 925, 1000 hPa. \\

\midrule

FengWu &
Pressure-level variables: geopotential (Z), specific humidity (Q), temperature (T), zonal wind (U), meridional wind (V).

Surface variables: mean sea-level pressure (MSL), 10m zonal wind (U10), 10m meridional wind (V10), 2m temperature (T2M). &
Pressure levels:
50, 100, 150, 200, 250, 300, 400, 500, 600, 700, 850, 925, 1000 hPa. \\

\midrule

FuXi &
Pressure-level variables: geopotential (Z), temperature (T), zonal wind (U), meridional wind (V), relative humidity (R).

Surface variables: 2m temperature (T2M), 10m zonal wind (U10), 10m meridional wind (V10), mean sea-level pressure (MSL), 6-hour accumulated total precipitation (TP6H). &
Pressure levels:
50, 100, 150, 200, 250, 300, 400, 500, 600, 700, 850, 925, 1000 hPa. \\

\midrule

Stormer &
Pressure-level variables: geopotential (Z), specific humidity (Q), temperature (T), zonal wind (U), meridional wind (V).

Surface variables: mean sea-level pressure (MSL), 10m zonal wind (U10), 10m meridional wind (V10), 2m temperature (T2M). &
Pressure levels:
50, 100, 150, 200, 250, 300, 400, 500, 600, 700, 850, 925, 1000 hPa. \\

\midrule

NeuralGCM &
Pressure-level variables: ggeopotential (Z), specific humidity (Q), temperature (T), zonal wind (U), meridional wind (V), specific cloud ice water content (CIWC), specific cloud liquid water content (CLWC).

Surface forcings: sea-surface temperature (SST), sea-ice concentration (SICONC). &
Pressure levels:
1, 2, 3, 5, 7, 10, 20, 30, 50, 70, 100, 125, 150, 175, 200, 225, 250, 300, 350, 400, 450, 500, 550, 600, 650, 700, 750, 775, 800, 825, 850, 875, 900, 925, 950, 975, 1000 hPa. \\

\bottomrule
\end{tabular}
}
\caption{Input variables and vertical levels used by different baseline models during ERA5-based inference.}
\label{tab:baseline_input_variables}
\end{table}

\begin{table}[t]
\centering
\small  
\begin{tabular}{
  >{\raggedright\arraybackslash}p{2.6cm} 
  >{\raggedright\arraybackslash}p{4.4cm} 
  >{\raggedright\arraybackslash}p{4.4cm} 
  >{\raggedright\arraybackslash}p{4.2cm}
}
\toprule
\textbf{Model} & \textbf{ERA5 checkpoint} & \textbf{IFS checkpoint} & \textbf{Source} \\
\midrule

AIFS &
\path{aifs-single-mse-1.1.ckpt} &
\path{aifs-single-mse-1.1.ckpt} &
\url{https://huggingface.co/ecmwf/aifs-single-1.1} \\

\midrule

Aurora &
\path{aurora-0.25-pretrained.ckpt} &
\path{aurora-0.25-finetuned.ckpt} &
\url{https://huggingface.co/microsoft/aurora} \\

\midrule

GraphCast &
\path{GraphCast-ERA51979-2017-resolution0.25-}\newline 
\path{pressurelevels37-mesh2to6-precipitationinputandoutput.npz} &
--- &
\url{https://github.com/google-deepmind/graphcast} \\

\midrule

GraphCast (operational) &
--- &
\path{GraphCast_operational-ERA5-HRES1979-2021-}\newline 
\path{resolution0.25-pressurelevels13-mesh2to6-precipitationoutputonly.npz} &
\url{https://github.com/google-deepmind/graphcast} \\

\midrule

Pangu-Weather &
\path{pangu_weather_6.onnx} &
\path{pangu_weather_6.onnx} &
\url{https://github.com/198808xc/Pangu-Weather} \\

\midrule

FengWu &
\path{fengwu_v1.onnx} &
\path{fengwu_v2.onnx} &
\url{https://github.com/OpenEarthLab/FengWu} \\

\midrule

FuXi &
\path{short.onnx}, \path{medium.onnx},\newline 
\path{long.onnx} &
\path{short.onnx}, \path{medium.onnx},\newline 
\path{long.onnx} &
\url{https://github.com/tpys/FuXi} \\

\midrule

Stormer &
\path{stormer_1.40625_patch_size_2.ckpt} &
\path{stormer_1.40625_patch_size_2.ckpt} &
\url{https://github.com/tung-nd/stormer} \\

\midrule

NeuralGCM &
\path{models_v1_deterministic_0_7_deg.pkl} &
--- &
\url{https://github.com/google-research/neuralgcm} \\

\bottomrule
\end{tabular}
\caption{Model checkpoints used for ERA5-based and IFS-based inference in this benchmark.}
\label{tab:baseline_model_versions}
\end{table}

\section{Additional Experimental Details}
\subsection{Baseline Models}
\label{app:baselines} 

All baseline models are widely recognized as strong data-driven numerical weather prediction systems, spanning graph-based, transformer-based, and hybrid architectures. For a fair comparison, we follow the official implementation and recommended inference protocol of each model. \textbf{The specific model checkpoints utilized for both ERA5-based and IFS-based (operational analysis) inference in this benchmark are summarized in Table~\ref{tab:baseline_model_versions}.} We briefly introduce these baselines below.

\textbf{AIFS}~\cite{Lang2024AIFS} is the operational machine-learning forecasting system developed by ECMWF, which adopts a data-driven architecture trained on global reanalysis and operational analysis data to produce medium-range forecasts with competitive skill against numerical weather prediction systems. It is based on a graph neural network (GNN) encoder-decoder and a sliding-window transformer processor.

\textbf{Aurora}~\cite{bodnar2025aurora} is a large-scale foundation model for the atmosphere, trained on heterogeneous meteorological datasets and designed to support multiple forecasting tasks through a unified pretrained representation. 

\textbf{GraphCast}~\cite{lam2023graphcast} formulates global weather prediction as learning message passing on a multi-scale graph defined over the sphere, enabling efficient long-range information propagation while preserving the geometric structure of the Earth. 

\textbf{Pangu-Weather}~\cite{bi2023pangu} employs a three-dimensional Earth-specific transformer architecture to model global atmospheric dynamics across pressure levels and surface variables, and produces forecasts in an autoregressive manner. 

\textbf{FuXi}~\cite{chen2023fuxi} is a cascade-based data-driven forecasting framework that uses separate neural networks for different lead-time ranges, improving medium-range prediction by reducing error accumulation over autoregressive rollouts. 

\textbf{FengWu}~\cite{chen2025operational} addresses global medium-range weather forecasting from a multi-modal and multi-task learning perspective, jointly modeling atmospheric variables and optimizing predictions across multiple lead times. It further incorporates a replay-buffer mechanism to enhance autoregressive forecasting skill at extended lead times.

\textbf{Stormer}~\cite{nguyen2024stormer} is a transformer-based global weather model that emphasizes scalable spatiotemporal representation learning for atmospheric states and supports efficient autoregressive forecasting. 

\textbf{NeuralGCM}~\cite{Kochkov2023neuralgcm} combines differentiable general circulation modeling with learned neural parameterizations, representing a hybrid approach that integrates physical structure with data-driven components.

\subsection{Complete List of Variables}
\label{app:varlist}

Table~\ref{tab:baseline_input_variables} lists the input variables and vertical levels used by each baseline model during inference. 
The reported configuration corresponds to ERA5-based inference in our experiments. 
For operational analysis data, all baselines use the same input variables and levels as in the ERA5 setting, except GraphCast. 
GraphCast uses 37 pressure levels for ERA5 inference, namely 1, 2, 3, 5, 7, 10, 20, 30, 50, 70, 100, 125, 150, 175, 200, 225, 250, 300, 350, 400, 450, 500, 550, 600, 650, 700, 750, 775, 800, 825, 850, 875, 900, 925, 950, 975, and 1000 hPa, whereas it uses 13 pressure levels for operational analysis inference, namely 50, 100, 150, 200, 250, 300, 400, 500, 600, 700, 850, 925, and 1000 hPa.
The surface variables used by GraphCast are identical in both settings.

\subsection{Evaluation Metrics}
\label{app:metrics}

We evaluate forecasts using four latitude-weighted metrics, including root-mean-square error (RMSE), anomaly correlation coefficient (ACC), bias, and activity, which account for varying grid-cell areas across latitudes and provide globally representative measures of forecast skill. Let $\hat{x}_{t}^{v}(i,j)$ and $x_{t}^{v}(i,j)$ denote the prediction and reference for variable $v$ at lead time $t$ and grid point $(i,j)$. 
We assign each latitude row a normalized area weight
\begin{equation}
    w_i = \frac{N_{\mathrm{lat}}\cos(\phi_i)}
    {\sum_{k=1}^{N_{\mathrm{lat}}}\cos(\phi_k)},
\end{equation}
where $\phi_i$ is the latitude of row $i$. 
For simplicity, we use $\langle a \rangle_w = \frac{1}{N_{\mathrm{lat}}N_{\mathrm{lon}}}\sum_{i,j} w_i a(i,j)$ to denote the latitude-weighted spatial average.

The weighted RMSE measures the magnitude of forecast errors:
\begin{equation}
    \mathrm{RMSE}_{t}^{v}
    =
    \sqrt{
    \left\langle
    \left(\hat{x}_{t}^{v}-x_{t}^{v}\right)^2
    \right\rangle_w
    } .
\end{equation}

The weighted ACC measures the spatial correlation between forecast and reference anomalies relative to the climatological mean $\bar{x}^{v}$:
\begin{equation}
    \mathrm{ACC}_{t}^{v}
    =
    \frac{
    \left\langle
    \left(\hat{x}_{t}^{v}-\bar{x}^{v}\right)
    \left(x_{t}^{v}-\bar{x}^{v}\right)
    \right\rangle_w
    }
    {
    \sqrt{
    \left\langle
    \left(\hat{x}_{t}^{v}-\bar{x}^{v}\right)^2
    \right\rangle_w
    }
    \sqrt{
    \left\langle
    \left(x_{t}^{v}-\bar{x}^{v}\right)^2
    \right\rangle_w
    }
    } .
\end{equation}

The weighted bias measures the signed mean forecast error:
\begin{equation}
    \mathrm{Bias}_{t}^{v}
    =
    \left\langle
    \hat{x}_{t}^{v}-x_{t}^{v}
    \right\rangle_w .
\end{equation}

Finally, weighted activity measures the spatial variability of the predicted anomaly field:
\begin{equation}
    \mathrm{Activity}_{t}^{v}
    =
    \sqrt{
    \left\langle
    \left(
    \hat{x}_{t}^{v}-\bar{x}^{v}
    -
    \left\langle \hat{x}_{t}^{v}-\bar{x}^{v} \right\rangle_w
    \right)^2
    \right\rangle_w
    } .
\end{equation}

\paragraph{Power Spectra.} The zonal energy spectrum measures the distribution of spatial variance across different physical scales (wavenumbers). For a given variable $v$ at lead time $t$ and latitude row $i$, the discrete Fourier transform (DFT) along the longitudinal dimension $j$ is:
\begin{equation}
    F_{t,k}^{v}(i) = \frac{1}{N_{\mathrm{lon}}} \sum_{j=0}^{N_{\mathrm{lon}}-1} \hat{x}_t^v(i,j) e^{-i 2\pi k j / N_{\mathrm{lon}}},
\end{equation}
where $k$ is the zonal wavenumber. The energy spectrum is then computed as:
\begin{equation}
    S_{t,0}^{v}(i) = C_i |F_{t,0}^{v}(i)|^2, \qquad S_{t,k}^{v}(i) = 2 C_i |F_{t,k}^{v}(i)|^2 \quad \mbox{for } k > 0,
\end{equation}
where $C_i = 2\pi R \cos(\phi_i)$ is the physical circumference of latitude $\phi_i$ ($R$ is the Earth's radius). The factor of $2$ ensures energy conservation by accounting for negative frequencies. Finally, the representative spectrum for active weather systems is obtained by averaging $S_{t,k}^{v}(i)$ over the mid-latitudes ($30^\circ < |\phi_i| < 60^\circ$).

All metrics are computed independently for each evaluated variable and forecast lead time.

\paragraph{Temperature Extreme Event Definition and Matching.}
For temperature extremes, let $T_{y,t}(x)$ denote the 2m temperature at location $x$ and time $t$ in year $y$. We define the daily maximum and minimum 2m temperature as
\begin{equation}
T^{\max}_{y,c}(x)=\max_{t\in c} T_{y,t}(x), 
\qquad
T^{\min}_{y,c}(x)=\min_{t\in c} T_{y,t}(x),
\end{equation}
where $c$ denotes a calendar day and $t\in c$ indexes all time steps within that day. For each calendar day, we construct a 15-day moving climatological window $\mathcal{W}_c=\{c-7,\ldots,c+7\}$ using 12 years of historical data $\mathcal{Y}_{\mathrm{hist}}$. This 12-year baseline is adopted in place of the standard 30-year climatology to balance computational efficiency with the constraints of our available historical dataset. The heatwave and cold surge thresholds are defined as the 90th and 10th percentiles \cite{perkins2013measurement}, 
respectively:
\begin{equation}
\tau^{\mathrm{H}}_{c}(x)
=
Q_{0.9}\left(
\left\{T^{\max}_{y,c'}(x): y\in\mathcal{Y}_{\mathrm{hist}},\; c'\in\mathcal{W}_c\right\}
\right),
\end{equation}
\begin{equation}
\tau^{\mathrm{C}}_{c}(x)
=
Q_{0.1}\left(
\left\{T^{\min}_{y,c'}(x): y\in\mathcal{Y}_{\mathrm{hist}},\; c'\in\mathcal{W}_c\right\}
\right).
\end{equation}
A heat-wave event is identified when the daily maximum temperature exceeds the corresponding threshold for at least three consecutive days; analogously, a cold surge event is identified when the daily minimum temperature falls below the corresponding threshold for at least three consecutive days. Specifically, for a three-day interval starting at day $c$, the binary event labels are
\begin{equation}
E^{\mathrm{H}}_{y,c}(x)
=
\prod_{\delta=0}^{2}
\mathbf{1}\left[
T^{\max}_{y,c+\delta}(x) > \tau^{\mathrm{H}}_{c+\delta}(x)
\right],
\qquad
E^{\mathrm{C}}_{y,c}(x)
=
\prod_{\delta=0}^{2}
\mathbf{1}\left[
T^{\min}_{y,c+\delta}(x) < \tau^{\mathrm{C}}_{c+\delta}(x)
\right].
\end{equation}
We compute the above binary event labels for both forecasts and ground truth. 
To account for small temporal displacement errors in predicted events, we match predicted and ground-truth events using the intersection-over-union (IoU) criterion. 
For a predicted event $\hat{E}$ and a ground-truth event $E$, the IoU is defined as
\begin{equation}
\mathrm{IoU}(\hat{E},E)
=
\frac{|\hat{E}\cap E|}
{|\hat{E}\cup E|},
\end{equation}
where $|\hat{E}\cap E|$ and $|\hat{E}\cup E|$ denote the number of overlapping and union event days, respectively. 
A predicted event is considered a true positive if it can be matched to a ground-truth event with
\begin{equation}
\mathrm{IoU}(\hat{E},E) \ge \gamma,
\end{equation}
where $\gamma=0.5$ unless otherwise specified. 
Each ground-truth event is matched to at most one prediction. 
Unmatched predicted events are counted as false positives, and unmatched ground-truth events are counted as false negatives. 

\subsection{Implementation Details}
\label{hardware}

\begin{table}[t]
\centering
\small

\begin{tabular}{lccccccc}
\toprule
 & Pangu & Stormer & FengWu & FuXi & AIFS & GraphCast & NeuralGCM \\
\midrule
Total time (h) & 12.1 & 10.6 & 20.8 & 19.5 & 10.1 & 14.3 & 92.9 \\
GPU (MiB) & 12{,}387 & 15{,}141 & 17{,}507 & 5{,}815 & 20{,}465 & 73{,}753 & 73{,}777 \\
Params (M) & 276.7 & 468.8 & 427.4 & 1563.3 & 253.0 & 36.5 & 31.1 \\
\bottomrule
\end{tabular}
\caption{Efficiency comparison of seven data-driven weather forecasting models. All statistics are evaluated with batch size $=1$, including total inference time, GPU memory consumption, and the number of model parameters.}
\label{tab:nwp_efficiency}
\end{table}

We implement all inference baselines in a unified evaluation pipeline and report efficiency under a fixed operational protocol. Specifically, we evaluate one full year (365 days), with 4 initialization times per day, and 40 forecast updates per initialization (6-hour cadence), resulting in 58,400 forecast updates per model for the sequential-throughput estimate. Unless otherwise stated, efficiency is measured with batch size set to 1, and we report parameter count, per-step inference latency (after warmup), peak GPU memory, and estimated total sequential GPU hours for the full protocol.

For NeuralGCM, we follow its native temporal integration setting and account for its 1-hour internal timestep when estimating long-horizon runtime, so that comparisons are aligned by forecast horizon rather than by internal step count. All experiments are conducted on our production compute cluster with multi-GPU nodes; each model is benchmarked on a single GPU process at a time, and multi-GPU parallelism is used only for throughput scaling across independent runs. The complete efficiency results are summarized in Table~\ref{tab:nwp_efficiency}.

Experiments are conducted on a server with 2× Intel Xeon Platinum 8457C CPUs and 8× NVIDIA H20 GPUs (97,871 MiB per GPU, driver 535.161.08).

\subsection{More Experimental Results}
\label{app:more_result}

\begin{figure}[t]
	\centering
	\includegraphics[width=\linewidth]{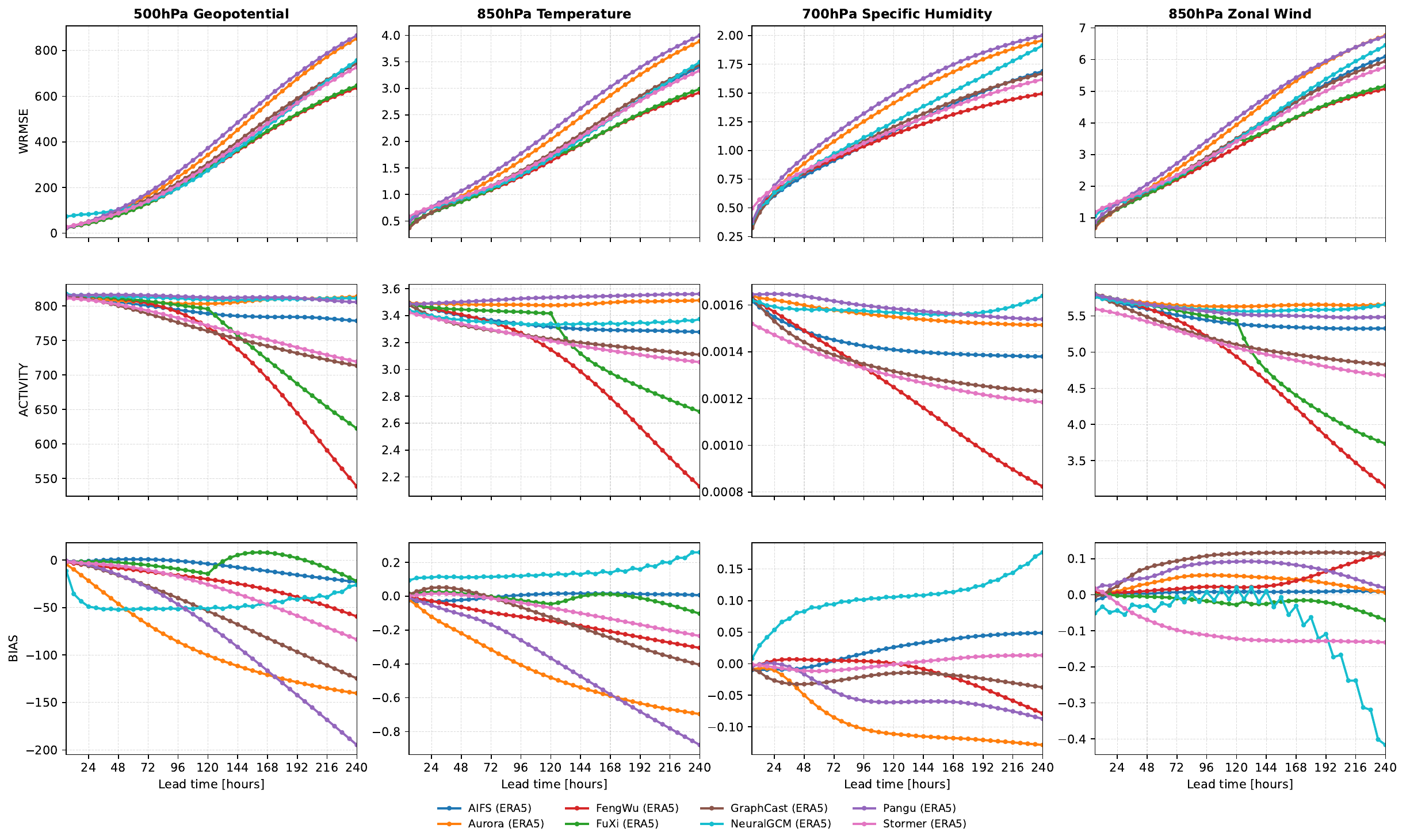}
	\caption{WRMSE $\downarrow$, Activity $\uparrow$, and Bias comparisons of eight baseline models for upper-air variables across ERA5 evaluations in global weather forecasting.}
\label{fig:era5_upper}
\end{figure}

\begin{figure}[t]
	\centering
	\includegraphics[width=\linewidth]{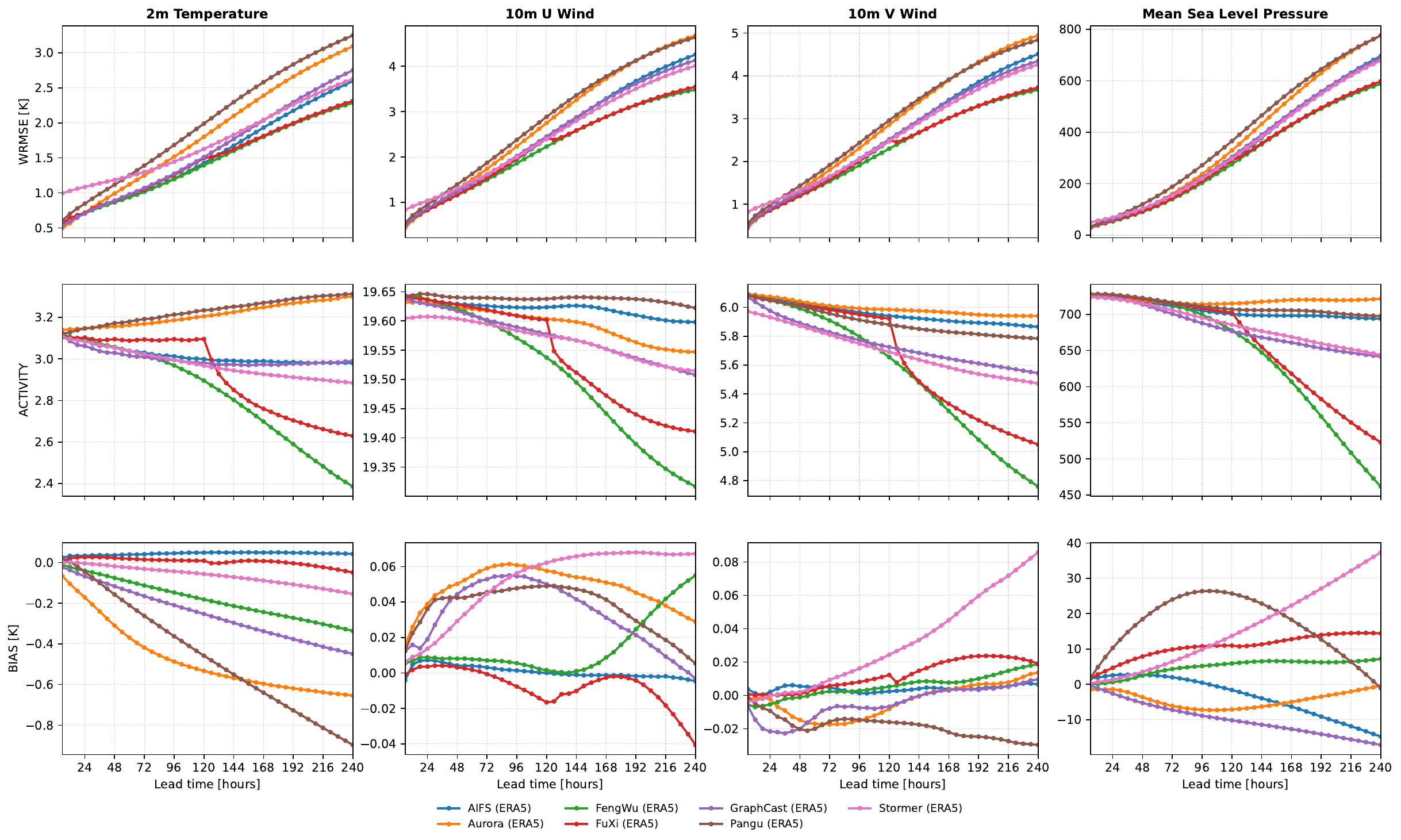}
	\caption{WRMSE $\downarrow$, Activity $\uparrow$, and Bias comparisons of eight baseline models for surface variables across ERA5 evaluations in global weather forecasting.}
\label{fig:era5_surface}
\end{figure}

\begin{figure}[t]
	\centering
	\includegraphics[width=\linewidth]{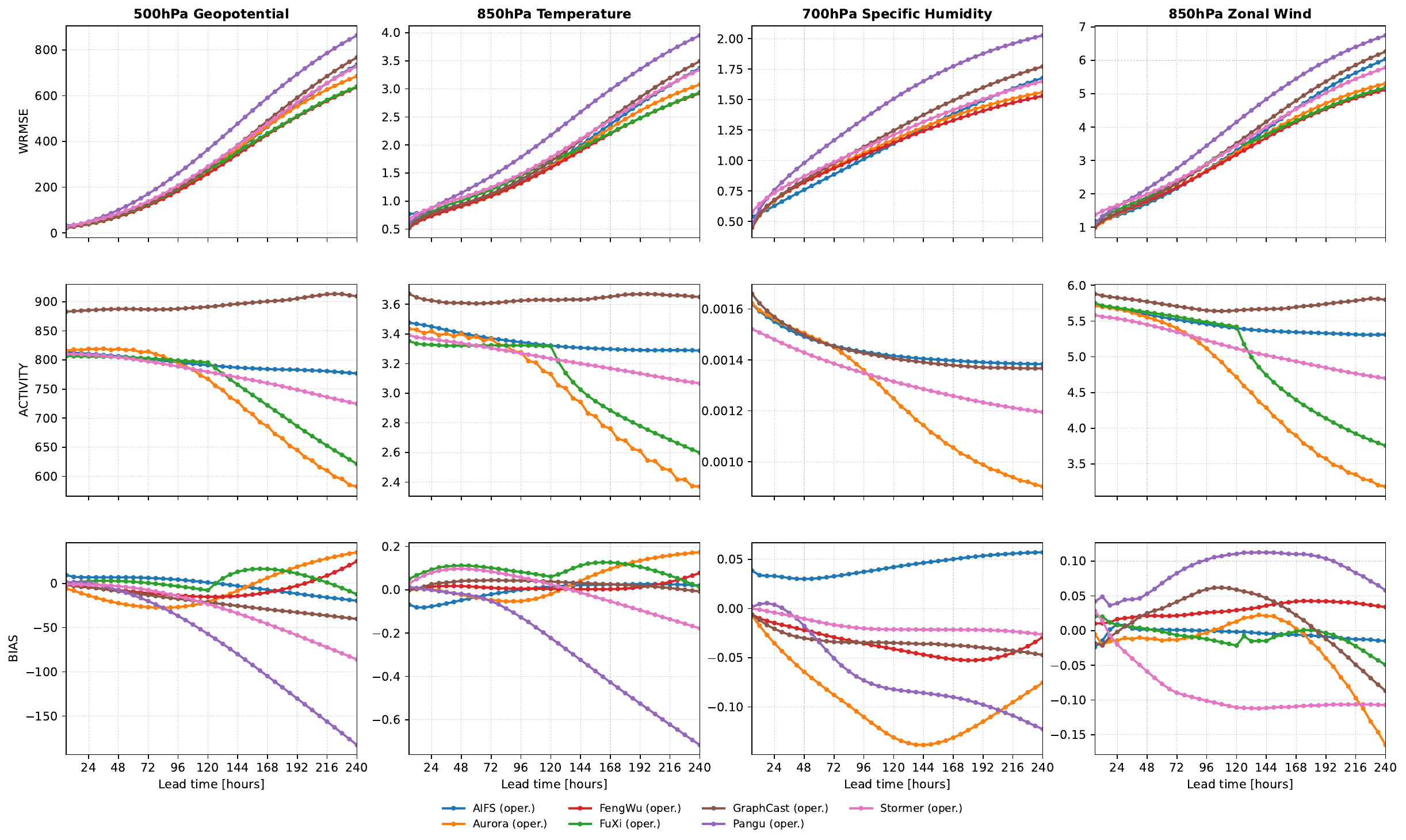}
	\caption{WRMSE $\downarrow$, Activity $\uparrow$, and Bias comparisons of eight baseline models for upper-air variables across operational analysis data evaluations in global weather forecasting.}
\label{fig:ifs_upper}
\end{figure}

\begin{figure}[t]
	\centering
	\includegraphics[width=\linewidth]{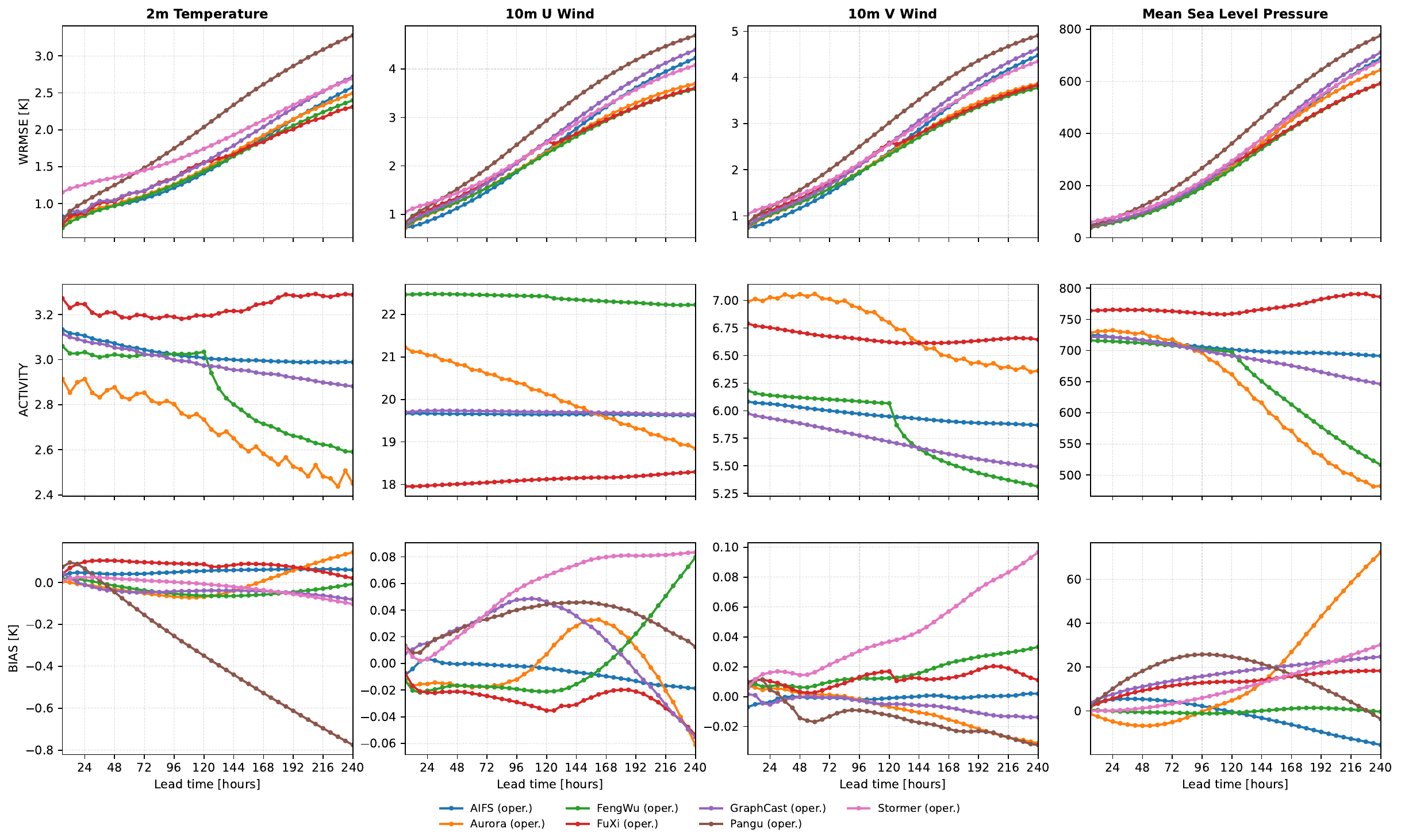}
	\caption{WRMSE $\downarrow$, Activity $\uparrow$, and Bias comparisons of eight baseline models for surface variables across operational analysis data evaluations in global weather forecasting.}
\label{fig:ifs_surface}
\end{figure}

\begin{figure}[t]
	\centering
	\includegraphics[width=\linewidth]{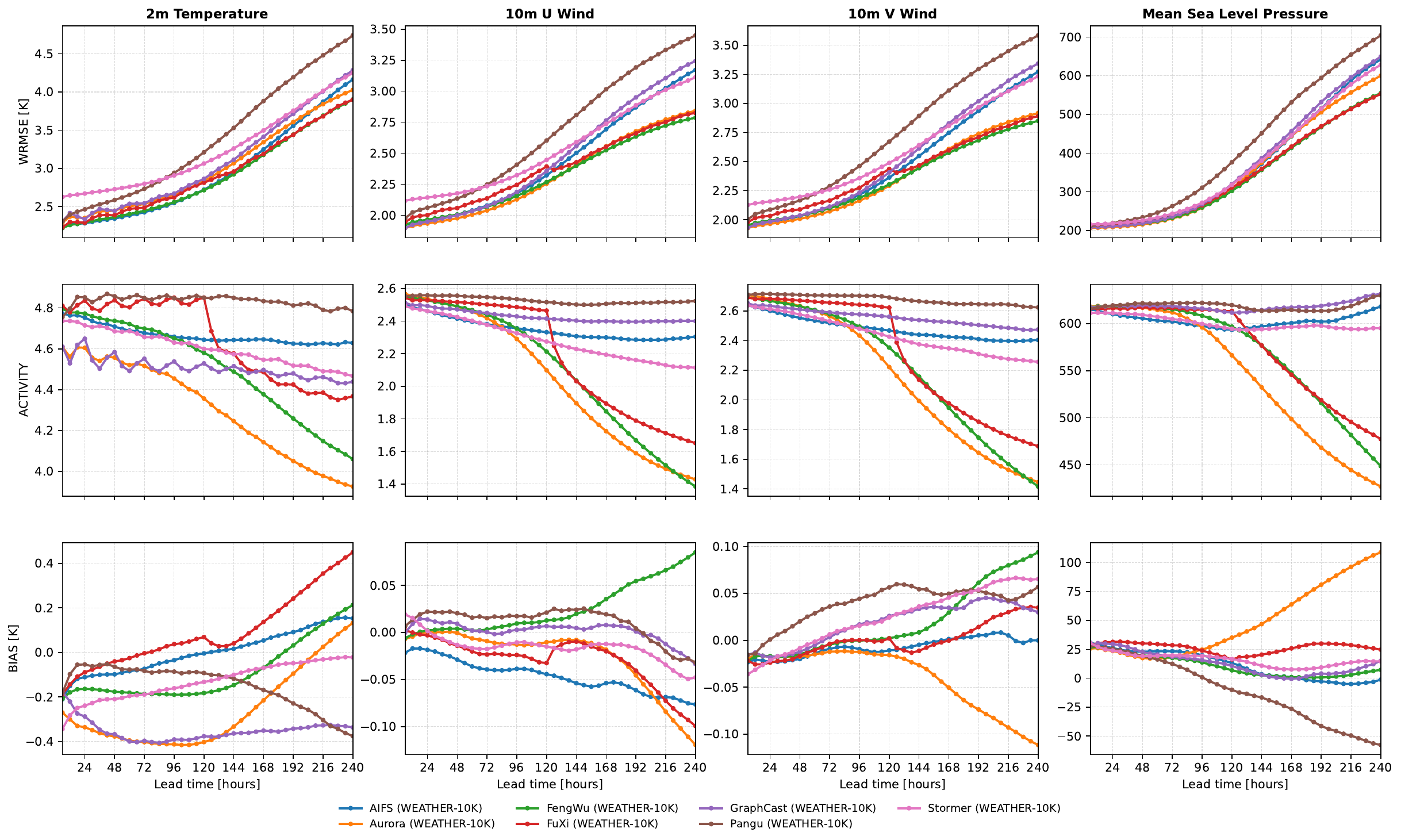}
	\caption{WRMSE $\downarrow$, Activity $\uparrow$, and Bias comparisons of eight baseline models for surface variables across WEATHER-10K evaluations in global weather forecasting.}
\label{fig:station_surface}
\end{figure}

\paragraph{Additional Overall Evaluation Results.} Figures~\ref{fig:era5_upper},~\ref{fig:era5_surface},~\ref{fig:ifs_upper},~\ref{fig:ifs_surface}, and~\ref{fig:station_surface} present the evolution of WRMSE, Bias, and Activity as functions of forecast lead time for different data-driven weather forecasting models. The metrics are evaluated on ERA5 reanalysis data, operational analysis data, and WEATHER-10K station observations, providing a comprehensive comparison of forecast accuracy, systematic bias, and variability preservation across multiple evaluation datasets.

\begin{figure}[t]
	\centering
    \includegraphics[width=\linewidth]{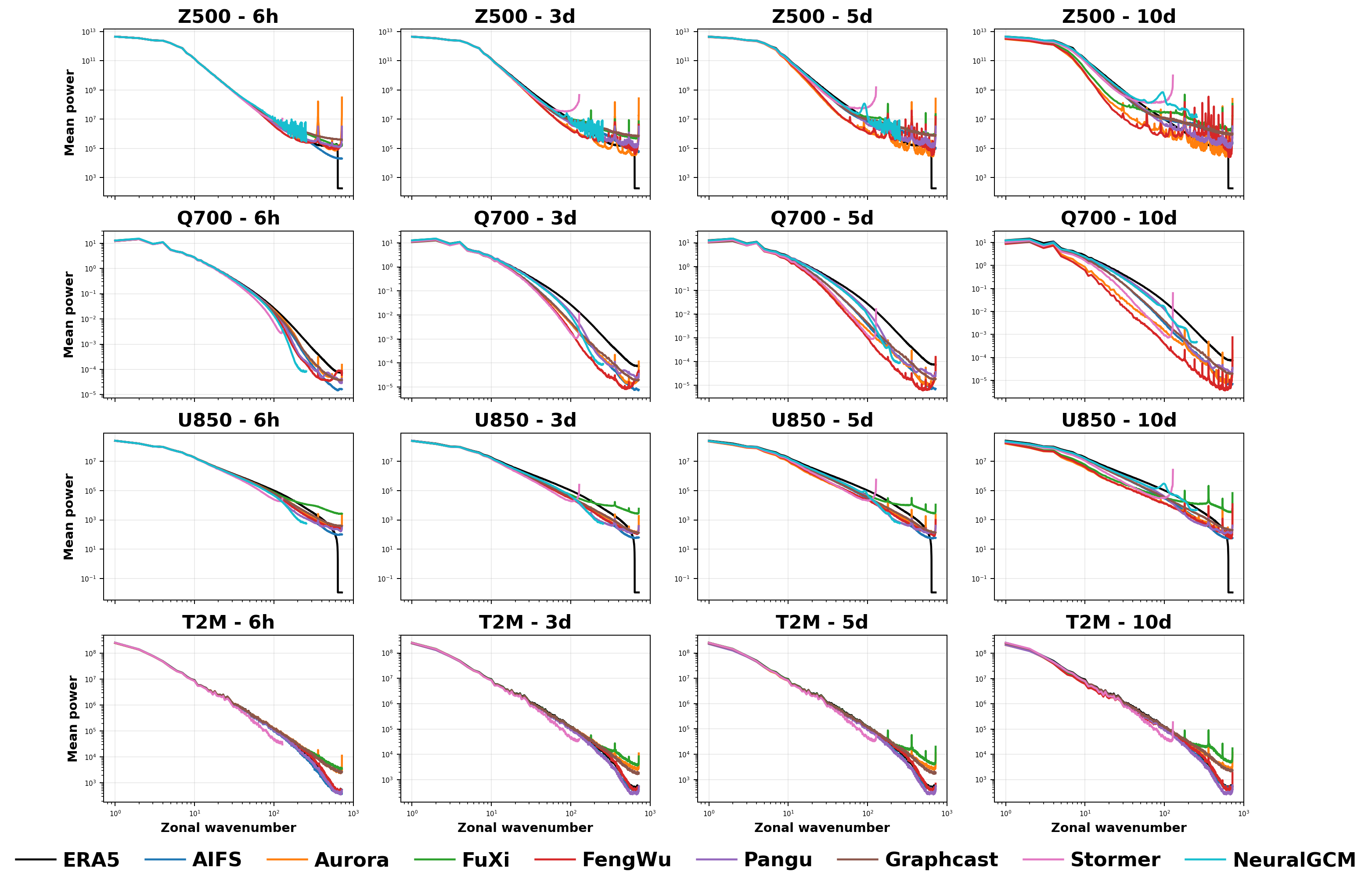}
	\caption{Zonal-mean power spectra for 500 hPa geopotential, 700 hPa specific humidity,
  850 hPa zonal wind, and 2 m temperature at leads of 6~h, 3~d, 5~d, and 10~d.
  Multiple AI models are compared against ERA5 (black)}
        \label{fig:power_spectra}
\end{figure}

Figure~\ref{fig:power_spectra} shows zonal-mean power spectra for four variables
(Z500, Q700, U850, and T2M) at lead times of 6~h, 3~d, 5~d, and 10~d.
The ERA5 curve is invariant across lead times and serves as a common reference.
Overall, most models are closer to ERA5 at low wavenumbers (large scales), while
larger deviations appear at high wavenumbers (small scales). From 6~h to 10~d,
high-wavenumber deviations increase for several variables. By variable, Q700 and
T2M show more pronounced differences at the high-wavenumber end, U850 remains
more consistent in the low-to-mid wavenumber range, and Z500 exhibits larger
inter-model spread in the spectral tail. Distinct local peaks and wiggles are also
visible at high wavenumbers for some models. NeuralGCM is reported only for upper-air
variables (Z500/Q700/U850), and is not shown for T2M.

\begin{figure}[t]
	\centering
	\includegraphics[width=\linewidth]{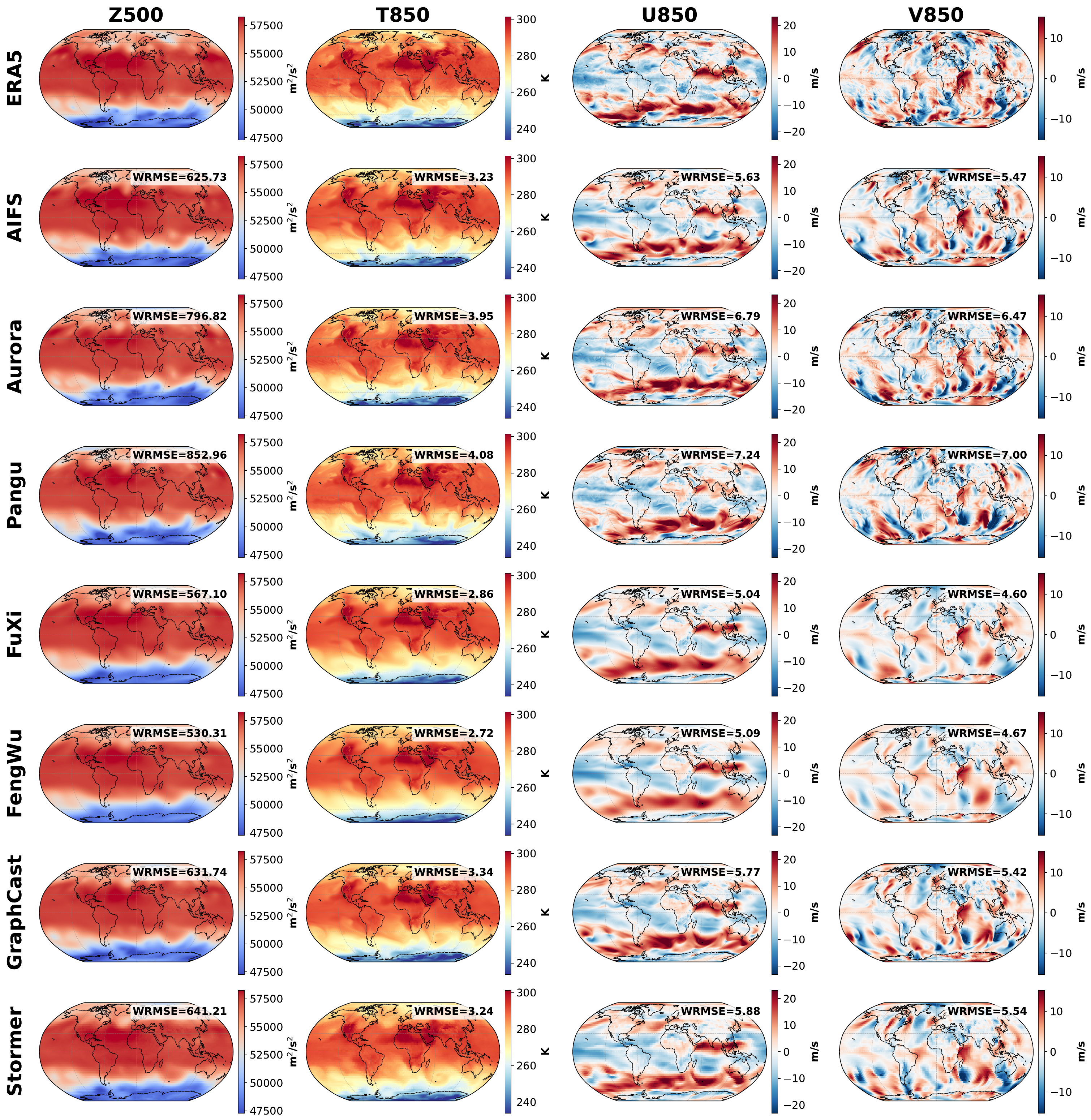}
	\caption{Visualization of 10-day global forecasts for upper-air variables generated by data-driven weather forecasting models, initialized at 00:00 UTC on July 10, 2025, with inference performed on ERA5.}
\label{fig:era5_global_upper}
\end{figure}

\begin{figure}[t]
	\centering
	\includegraphics[width=\linewidth]{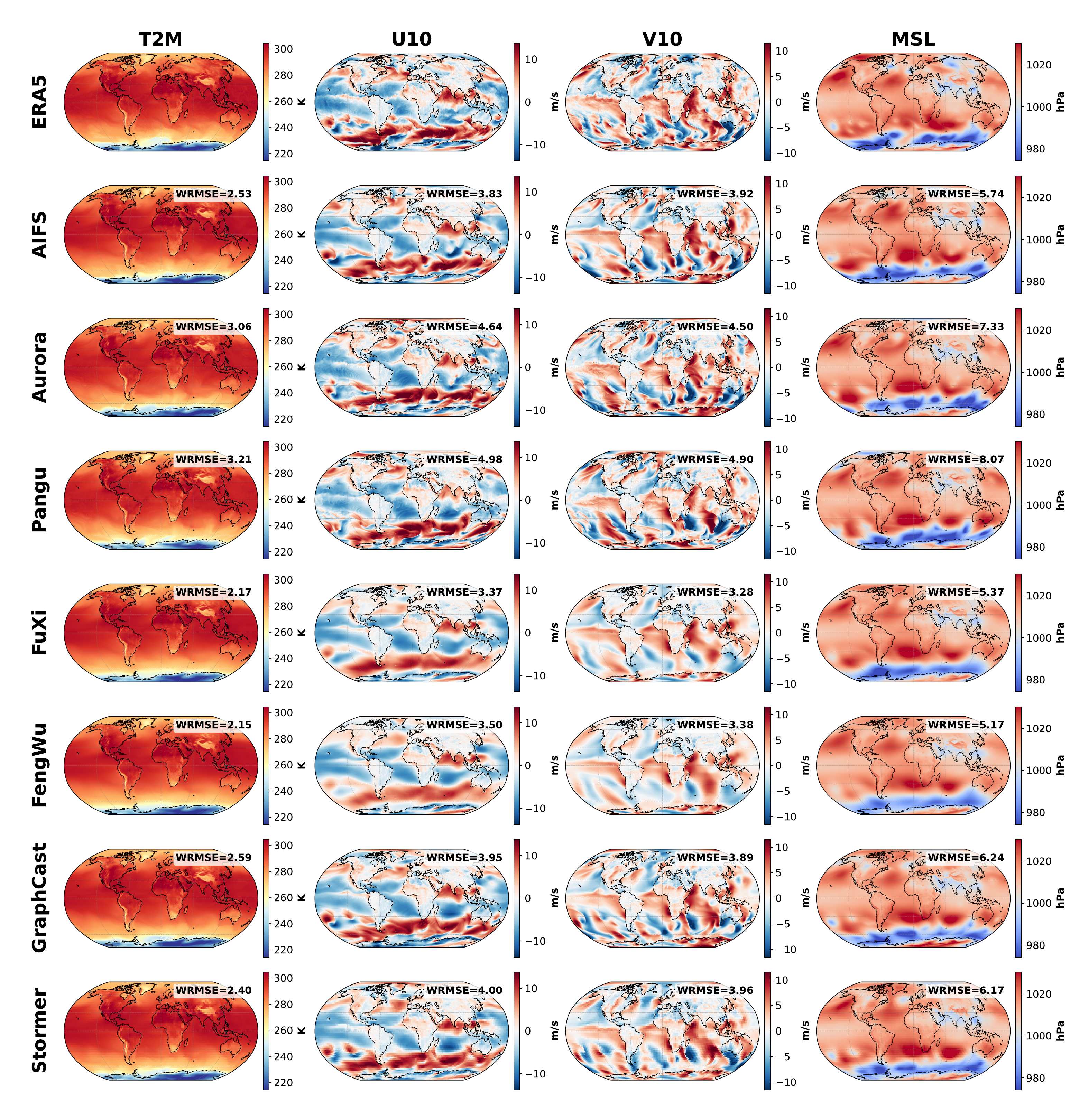}
	\caption{Visualization of 10-day global forecasts for surface variables generated by data-driven weather forecasting models, initialized at 00:00 UTC on July 10, 2025, with inference performed on ERA5.}
\label{fig:era5_global_surface}
\end{figure}

\begin{figure}[t]
	\centering
	\includegraphics[width=\linewidth]{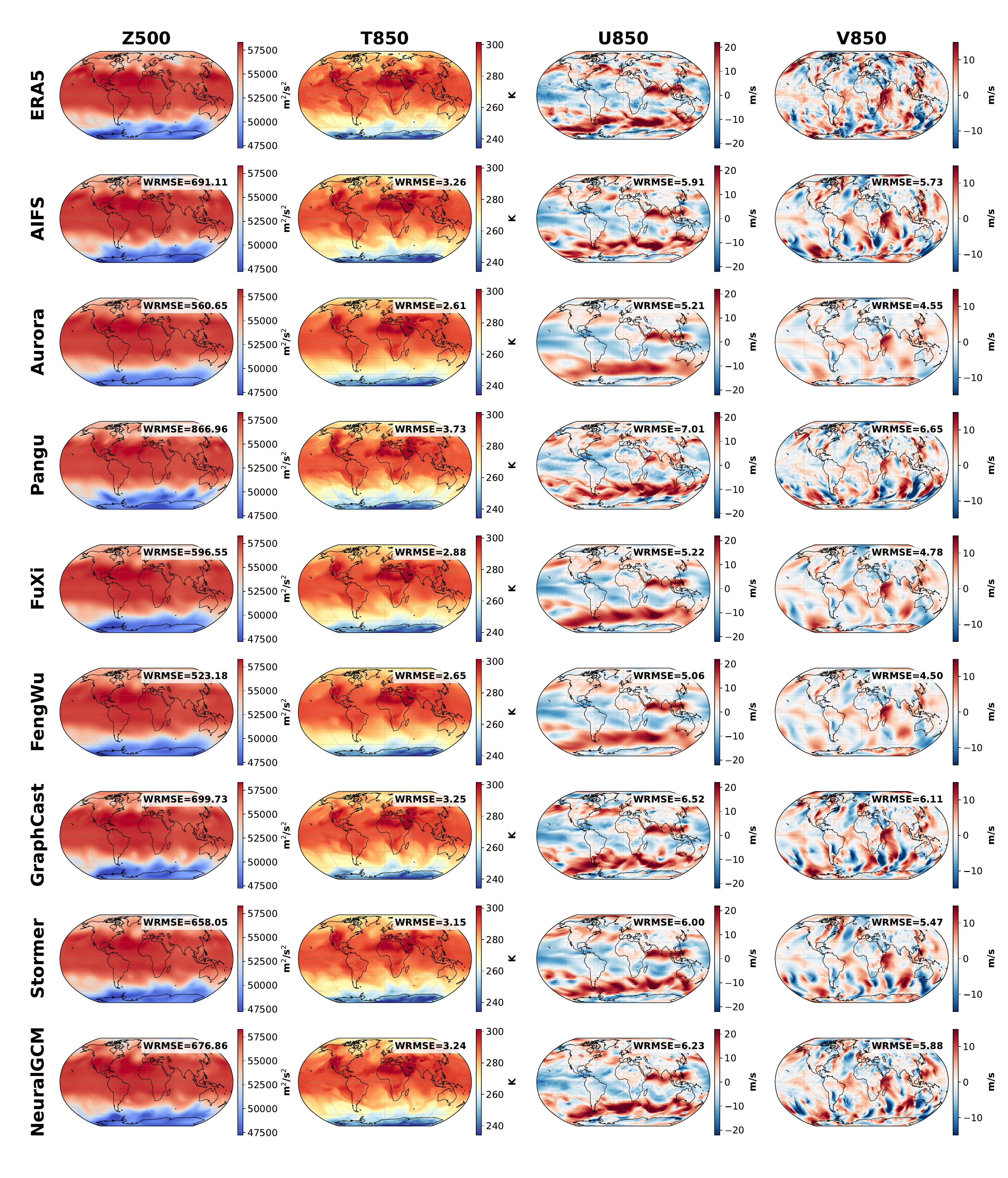}
	\caption{Visualization of 10-day global forecasts for upper-air variables generated by data-driven weather forecasting models, initialized at 00:00 UTC on July 10, 2025, with inference performed on operational analysis data.}
\label{fig:ifs_global_upper}
\end{figure}

\begin{figure}[t]
	\centering
	\includegraphics[width=\linewidth]{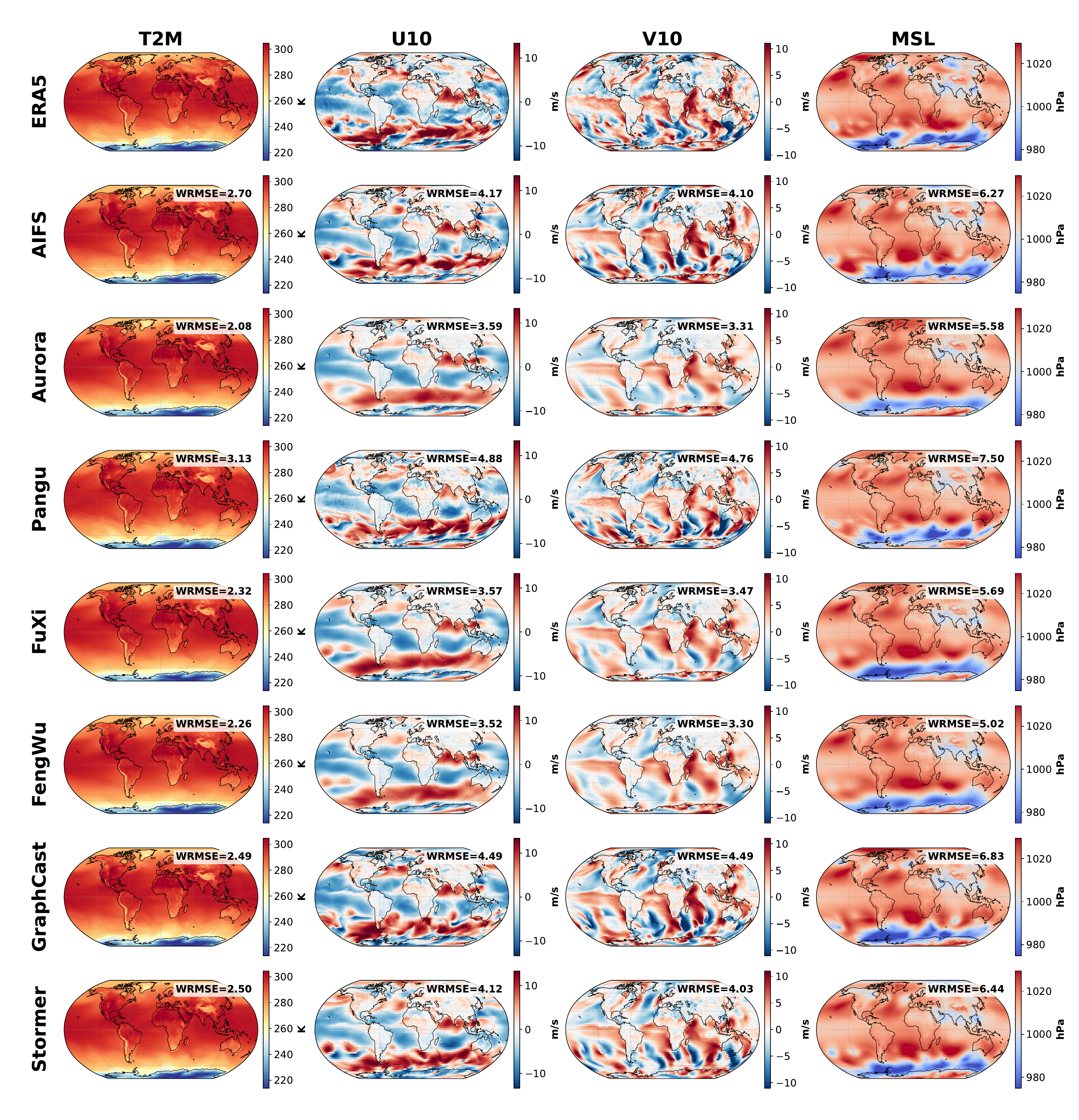}
	\caption{Visualization of 10-day global forecasts for surface variables generated by data-driven weather forecasting models, initialized at 00:00 UTC on July 10, 2025, with inference performed on operational analysis data.}
\label{fig:ifs_global_surface}
\end{figure}

\paragraph{Global Visualizations.} Figures~\ref{fig:era5_global_upper},~\ref{fig:era5_global_surface},~\ref{fig:ifs_global_upper}, and~\ref{fig:ifs_global_surface} show global visualizations of near-surface and upper-air forecasts produced by data-driven weather forecasting models. The forecasts are initialized from ERA5 and operational analysis data at 00:00 UTC on July 10, 2025, and the results at a lead time of 10 days are presented to illustrate the large-scale spatial structures and forecast behaviors across different atmospheric variables.

\paragraph{Additional Tropical Cyclone Results.}
\begin{figure}[t]
	\centering
    \includegraphics[width=\linewidth]{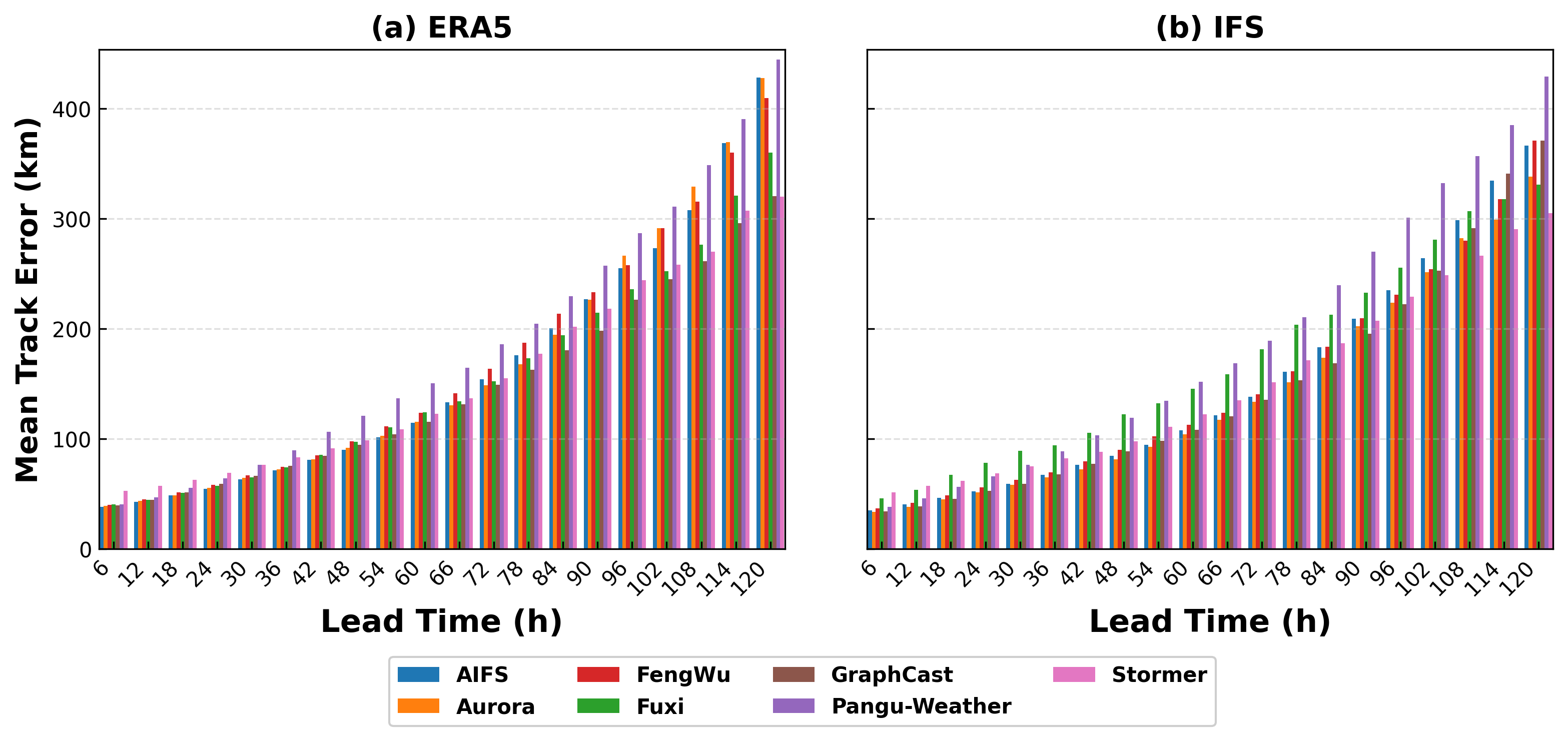}
	\caption{Mean tropical cyclone track error (km) versus lead time (6--120 h) from June to December 2025 for seven AI weather models, comparing ERA5-initialized and IFS-initialized forecasts.}
    \label{fig:tc_2}
\end{figure}
Figure~\ref{fig:tc_2} reports lead-time-dependent track errors for all seven models under ERA5 and IFS initialization. For all models, errors increase with lead time. Differences among models are small at short leads and become larger at medium-to-long leads. The relative ordering is broadly similar across the two initialization settings, while the error spread is more visible after longer lead times.




\begin{figure}[t]
	\centering
    \includegraphics[width=\linewidth]{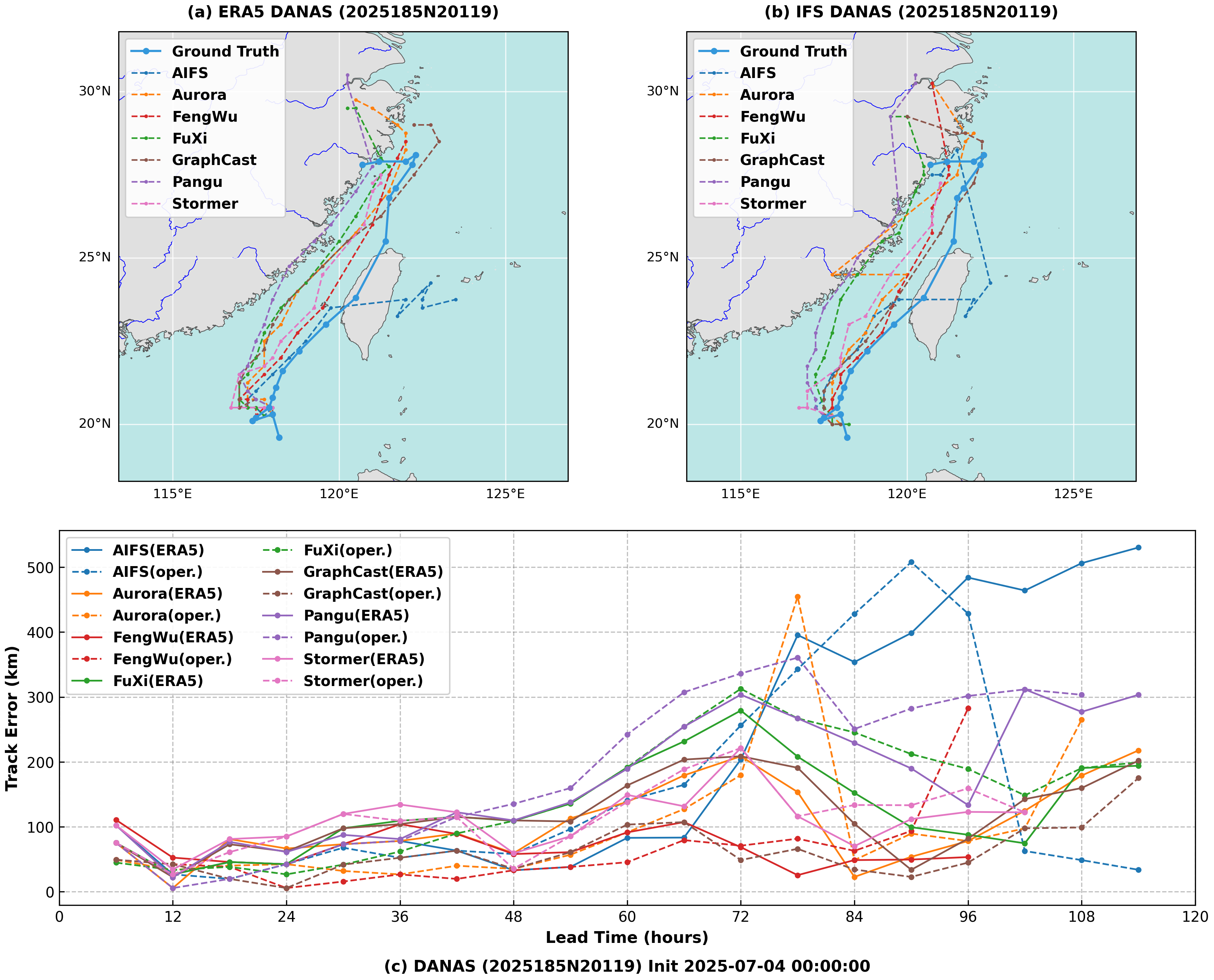}
	\caption{Track forecasts initialized from (a) ERA5 and (b) IFS analysis for Typhoon Danas (IBTrACS ID 2025185N20119; initialization 0000 UTC 4 July 2025). Observed track from IBTrACS (blue). (c) Mean-distance track error versus lead time (0--120 h) for each model and initialization.}
        \label{fig:tc_case_Danas}
\end{figure}

\begin{figure}[t]
	\centering
    \includegraphics[width=\linewidth]{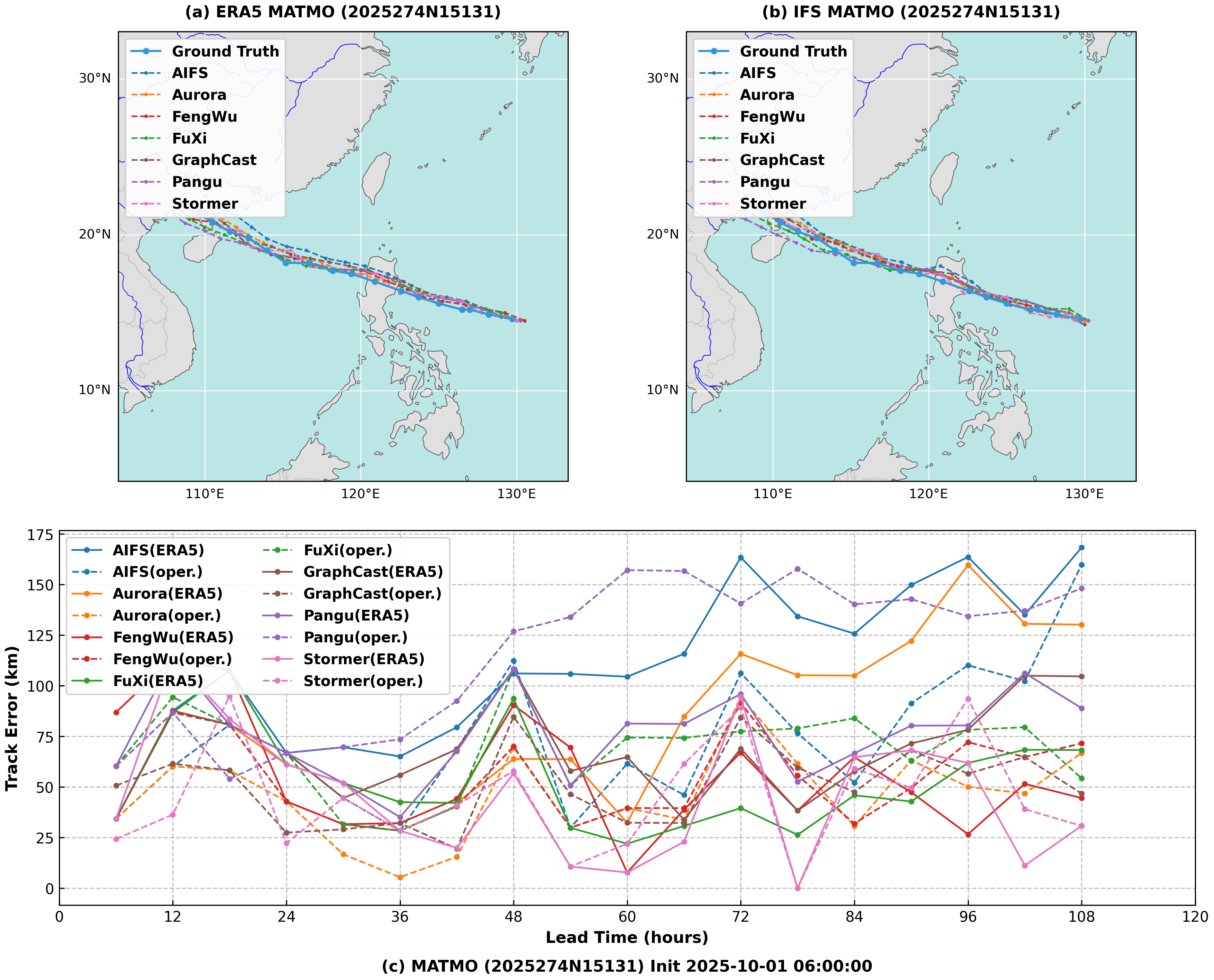}
	\caption{Track forecasts initialized from (a) ERA5 and (b) IFS analysis for Typhoon Matmo (IBTrACS ID 2025274N15131; initialization 0600 UTC 1 October 2025). Observed track from IBTrACS (blue). (c) Mean-distance track error versus lead time (0--120 h) for each model and initialization.}
        \label{fig:tc_case_MATMO}
\end{figure}

\begin{figure}[t]
	\centering
    \includegraphics[width=\linewidth]{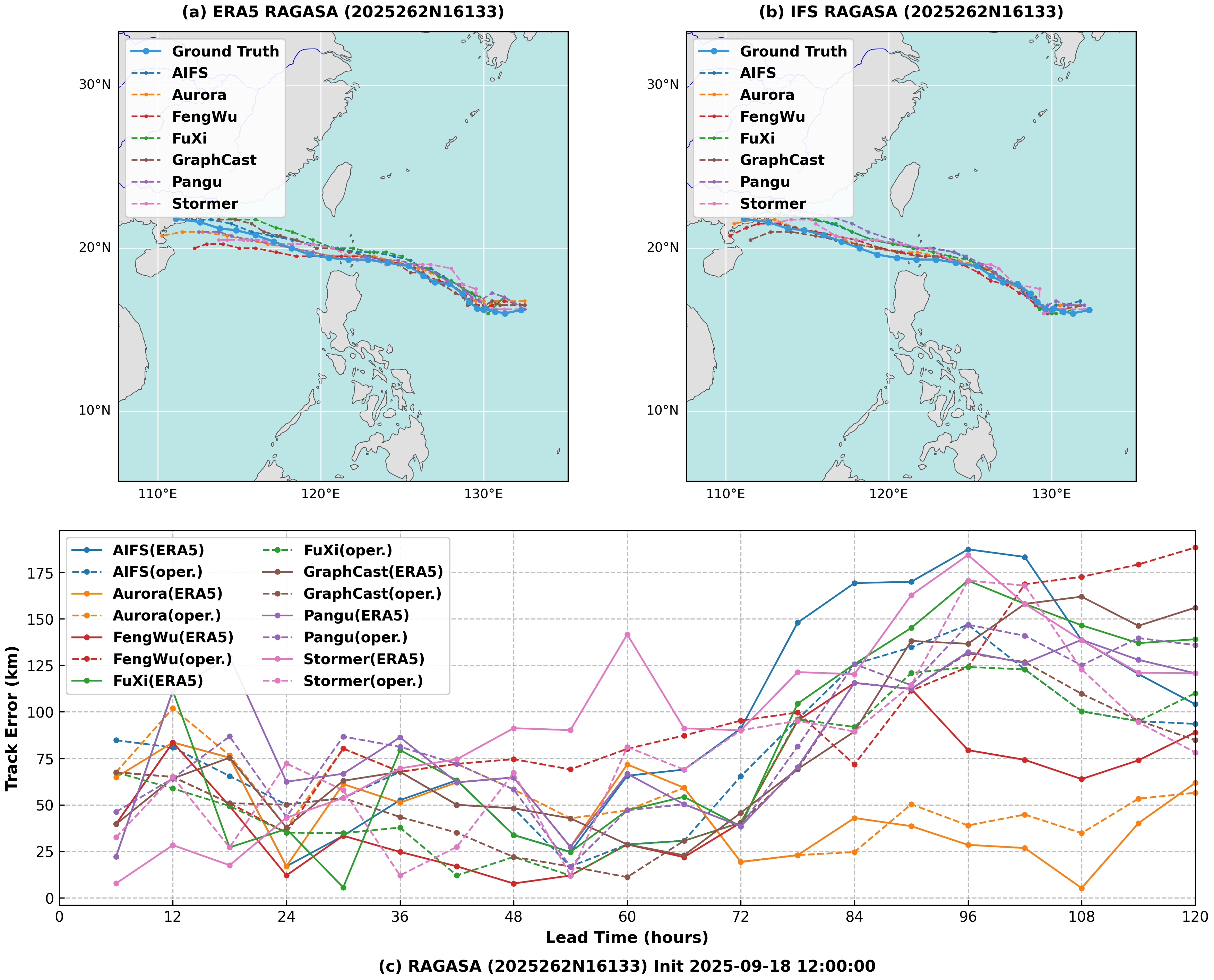}
	\caption{Track forecasts initialized from (a) ERA5 and (b) IFS analysis for Typhoon Ragasa (IBTrACS ID 2025262N16133; initialization 1200 UTC 18 September 2025). Observed track from IBTrACS (blue). (c) Mean-distance track error versus lead time (0--120 h) for each model and initialization.}
        \label{fig:tc_case_RAGASA}
\end{figure}


Figures~\ref{fig:tc_case_Danas},~\ref{fig:tc_case_MATMO}, and~\ref{fig:tc_case_RAGASA} provide three representative event-level tropical cyclone evaluations (Danas, Matmo, and Ragasa) under a unified protocol. For each event, panels (a) and (b) show forecast tracks initialized from ERA5 and IFS, respectively, against the IBTrACS reference track, and panel (c) reports model-wise track error as a function of lead time (0--120 h). Across the three cases, near-initial lead errors are generally smaller and become larger at later leads, while the spread among models is limited at short leads and more pronounced at medium-to-long leads. The degree of spread and the ordering among models vary by event, indicating clear case dependence in track behavior. Overall, these case studies are consistent with the aggregate statistics in the main benchmark and serve as supplementary, event-wise evidence of inter-model differences under the same data window and evaluation settings (June to December 2025).


\begin{figure}[t]
	\centering
    \includegraphics[width=\linewidth]{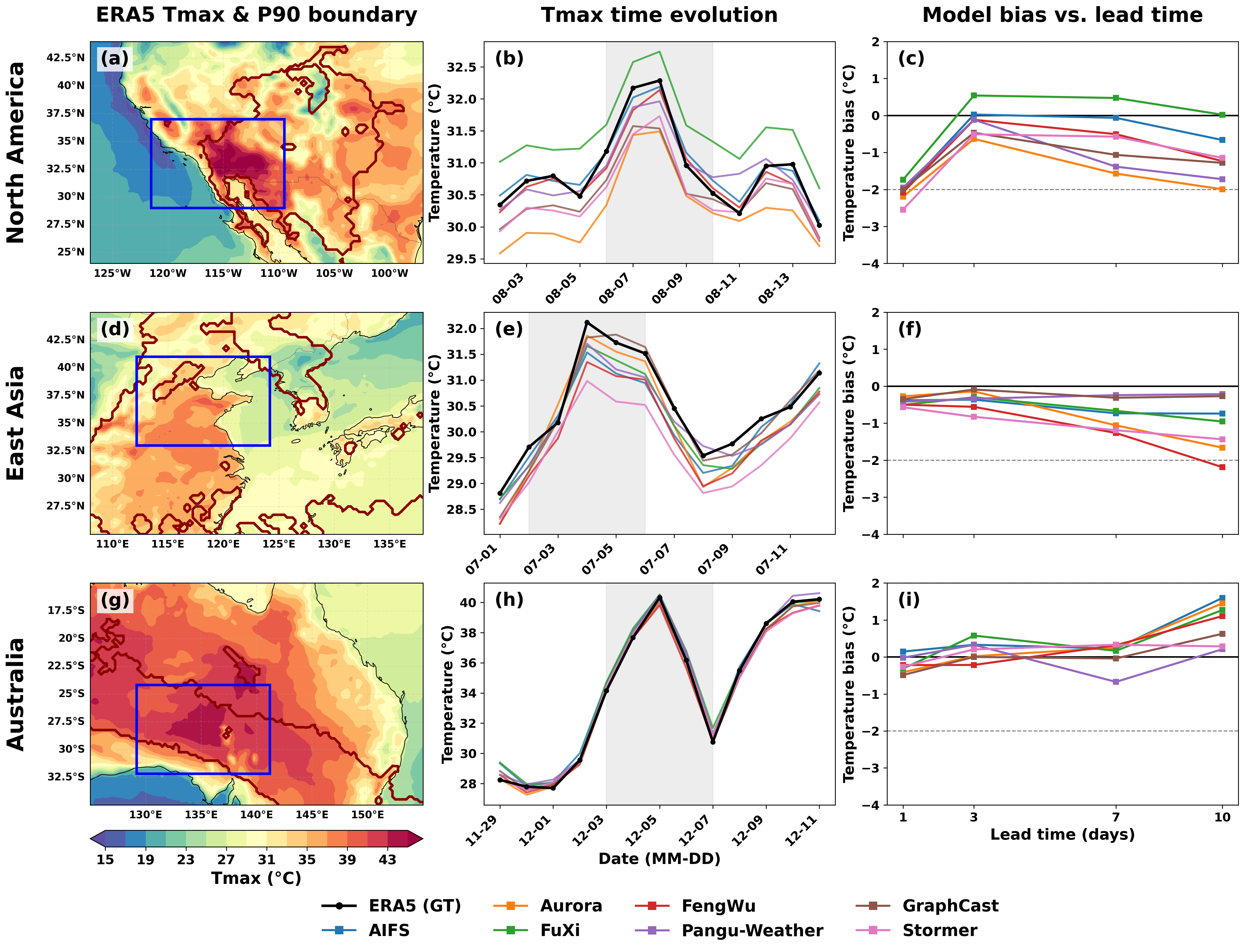}
		\caption{Multiple AI models for three major heatwave (HW) events in the second half of 2025: (a--c) North America; (d--f) East Asia; and (g--i) Australia. \textbf{(a, d, and g)} ERA5 daily maximum temperature (Tmax, shading, in $^\circ$C) on the peak day of each event. Dark red contours denote the object-based HW boundaries (Tmax exceeding the local 90th percentile climatological threshold), and blue boxes indicate the regions for area-averaged analysis. \textbf{(b, e, and h)} Temporal evolution of the area-weighted Tmax over the blue boxes for ERA5 (black) and seven AI models at a 3-day forecast lead time. The core HW period is shaded in gray, with vertical dashed lines marking the peak day. \textbf{(c, f, and i)} Model bias in Tmax ($^\circ$C) as a function of forecast lead time (1, 3, 7, and 10 days), averaged over the core HW period. Negative values at longer lead times reveal a systematic underestimation of extreme heat intensity.}
        \label{fig:heatwave}
\end{figure}


\begin{figure}[t]
	\centering
    \includegraphics[width=\linewidth]{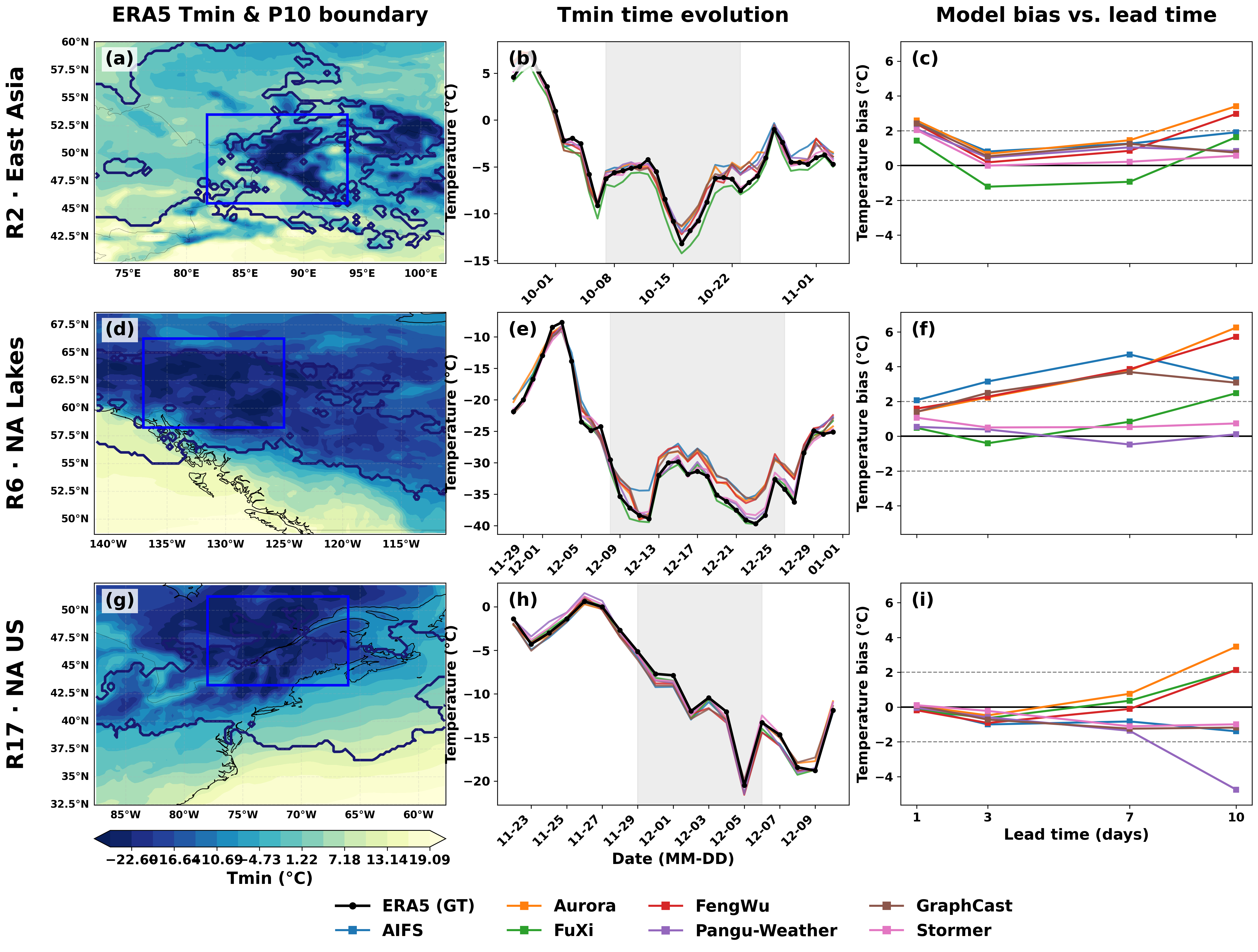}
	\caption{Multiple AI models for three major cold-surge (CS) events in the second half of 2025: (a--c) East Asia; (d--f) North America (Great Lakes); and (g--i) North America (CONUS). \textbf{(a, d, and g)} ERA5 daily minimum temperature (Tmin, shading, in $^\circ$C) on the peak day of each event. Dark blue contours denote the object-based CS boundaries (Tmin below the local 10th percentile climatological threshold), and blue boxes indicate the regions for area-averaged analysis. \textbf{(b, e, and h)} Temporal evolution of the area-weighted Tmin over the blue boxes for ERA5 (black) and seven AI models at a 3-day forecast lead time. The core CS period is shaded in gray, with vertical dashed lines marking the peak day. \textbf{(c, f, and i)} Model bias in Tmin ($^\circ$C) as a function of forecast lead time (1, 3, 7, and 10 days), averaged over the core CS period. Positive values at longer lead times indicate a systematic warm bias (underestimation of cold extremes).}
        \label{fig:coldwave}
\end{figure}

\paragraph{Additional Case-Based Results for Extreme Temperature Events.}

Figures~\ref{fig:heatwave} and~\ref{fig:coldwave} report case-based results for three heatwave events and three cold-surge events in the second half of 2025. Each figure uses the same three-panel layout: (i) peak-day ERA5 temperature field with an object-based event boundary and analysis box, (ii) area-averaged temperature time series at 3-day lead, and (iii) lead-time-dependent bias (1, 3, 7, and 10 days) averaged over the event core period.

From the heatwave cases, two consistent observations are visible. First, agreement with ERA5 is generally higher at short lead times than at longer lead times. Second, several models show increasingly negative Tmax bias at longer leads in the strongest events, indicating reduced peak-intensity fidelity as lead time increases.

From the cold-surge cases, a parallel pattern is observed. Model spread increases with lead time, and positive Tmin bias becomes more common at longer leads, corresponding to weaker simulated cold intensity relative to ERA5. The magnitude of this behavior is case-dependent across regions.


\end{document}